\documentclass{article}

\usepackage[nonatbib,preprint]{arxiv}

\usepackage[utf8]{inputenc}
\usepackage[ruled]{algorithm2e}
\usepackage{algcompatible}
\usepackage[intlimits]{amsmath}
\usepackage{titlesec}
\usepackage{float}
\usepackage{array}
\usepackage{multirow}
\usepackage[ngerman,english]{babel}
\usepackage{graphicx,wrapfig}
\usepackage[toc]{appendix}
\usepackage{etaremune}
\usepackage[inline]{enumitem}
\usepackage[onehalfspacing]{setspace}
\usepackage[T1]{fontenc}
\usepackage{amsfonts}
\usepackage{amsthm,amssymb}
\usepackage{mathtools}
\usepackage[most]{tcolorbox}
\usepackage{upgreek}
\usepackage[format=hang,font=small]{caption}
\usepackage[colorlinks,linkcolor=black,citecolor=black,urlcolor=black,breaklinks]{hyperref}
\usepackage{url}
\usepackage{booktabs}
\usepackage{nicefrac}
\usepackage{microtype}
\usepackage{fancyhdr}
\usepackage{cleveref}
\usepackage{adjustbox}
\usepackage{perpage}
\usepackage{doi}
\graphicspath{{Figs/}}
\MakePerPage{footnote}
\DeclareMathOperator*{\argmin}{arg\,min}
\DeclareMathOperator*{\argmax}{arg\,max}

\DeclareMathOperator*{\median}{median}
\titlespacing{\section}{0pt}{5pt}{5pt}
\titlespacing{\subsection}{0pt}{5pt}{5pt}
\titlespacing{\subsubsection}{0pt}{5pt}{5pt}
\title{Hierarchical Multiresolution Feature- and Prior-based Graphs for Classification}
\author{Faezeh Fallah\\Institute of Signal Processing and System Theory\\University of Stuttgart,~Pfaffenwaldring 47\\70569 Stuttgart,~Germany\\\texttt{faezeh.fallah@iss.uni-stuttgart.de}}
\begin{document}
\maketitle
\begin{abstract}
To incorporate spatial (neighborhood) and bidirectional hierarchical relationships as well as features and priors of the samples into their classification, we formulated the classification problem on three variants of multiresolution neighborhood graphs and the graph of a hierarchical conditional random field. Each of these graphs was weighted and undirected and could thus incorporate the spatial or hierarchical relationships in all directions. In addition, each variant of the proposed neighborhood graphs was composed of a spatial feature-based subgraph and an aspatial prior-based subgraph. It expanded on a random walker graph by using novel mechanisms to derive the edge weights of its spatial feature-based subgraph. These mechanisms included implicit and explicit edge detection to enhance detection of weak boundaries between different classes in spatial domain. The implicit edge detection relied on the outlier detection capability of the Tukey's function and the classification reliabilities of the samples estimated by a hierarchical random forest classifier. Similar mechanism was used to derive the edge weights and thus the energy function of the hierarchical conditional random field. This way, the classification problem boiled down to a system of linear equations and a minimization of the energy function which could be done via fast and efficient techniques.
\end{abstract}
\section{Background and Motivation}
\label{sec:BackMotivGraphs}
A hierarchical classifier such as the random forest classifier, proposed in \cite{Fallah2018a,Fallah2018p,Fallah2019a,FallahJ2019}, could classify multiresolution validation/test samples ${\{\mathbb{D}_r=\mathbb{D}_{r,\mathrm{val}}\cup\mathbb{D}_{r,\mathrm{test}}\}}_{r=0}^{n_{\mathrm{lay}}}$ by using their squared features and their inter-resolution hierarchical relationships in coarse-to-fine (parent-to-child) direction. However, it did not consider the hierarchical relationships of the samples in a fine-to-coarse (child-to-parent) direction and ignored their intra-resolution spatial (neighborhood) relationships. Due to the regional continuity of an object's features in an image, the spatial relationships of samples were of particular importance for image segmentation. Also, consideration of the hierarchical relationships of the samples in both coarse-to-fine and fine-to-coarse directions could enhance the accuracy and the reliability of the classifications.

To incorporate the intra-resolution spatial relationships of the samples $\mathbb{D}_r=\mathbb{D}_{r,\mathrm{val}}\cup\mathbb{D}_{r,\mathrm{test}}$ of each resolution $r\in\{0,\cdots,n_{\mathrm{lay}}\}$ into their classifications, these relationships and other information of the samples got encoded into a \textbf{neighborhood graph} $\mathcal{G}_r$. This neighborhood graph was the weighted undirected graph of a conditional random field (CRF). It encoded the spatial (neighborhood) relationships, the features differences, and the prior classification probabilities of the samples $\mathbb{D}_r$ of each resolution $r\in\{0,\cdots,n_{\mathrm{lay}}\}$ into its edge weights (topology). Accordingly, it was denoted by $\mathcal{G}_r=\mathcal{G}_{r,\mathrm{feats}}\bigcup\mathcal{G}_{r,\mathrm{prior}}$ with $\mathcal{G}_{r,\mathrm{feats}}=(\mathbb{V}_{r,\mathrm{feats}},\mathbb{E}_{r,\mathrm{feats}})$ being a spatial feature-based subgraph and $\mathcal{G}_{r,\mathrm{prior}}=(\mathbb{V}_{r,\mathrm{prior}},\mathbb{E}_{r,\mathrm{prior}})$ being an aspatial prior-based subgraph. To classify the samples $\mathbb{D}_r$ over the graph $\mathcal{G}_r$ either some of them should be prelabeled or the classification priors of unlabeled samples should be provided. A prelabeled sample (vertex) was called a \textbf{seed}. For this, $\mathbb{D}_r$ got divided into the labeled $\mathbb{D}_{r,\mathrm{lab}}$ and unlabeled $\mathbb{D}_{r,\mathrm{unl}}$ subsets with $\mathbb{D}_{r,\mathrm{lab}}\cup\mathbb{D}_{r,\mathrm{unl}}=\mathbb{D}_r$ and $\mathbb{D}_{r,\mathrm{lab}}\cap\mathbb{D}_{r,\mathrm{unl}}=\emptyset$.

The spatial feature-based subgraph $\mathcal{G}_{r,\mathrm{feats}}=(\mathbb{V}_{r,\mathrm{feats}},\mathbb{E}_{r,\mathrm{feats}})$ encoded the spatial (neighborhood) relationships and the features differences of the samples $\mathbb{D}_r$ into its edge weights (topology). To this end, each sample $v_{r,j}\in\mathbb{D}_r$ was represented by a vertex $v_{r,j}\in\mathbb{V}_{r,\mathrm{feats}}$. Thus, $|\mathbb{V}_{r,\mathrm{feats}}|=|\mathbb{D}_r|$. If the samples $v_{r,j}\in\mathbb{D}_r$ and $v_{r,k}\in\mathbb{D}_r$ were direct spatial neighbors of each other, then $v_{r,j}\in\mathbb{V}_{r,\mathrm{feats}}$ and $v_{r,k}\in\mathbb{V}_{r,\mathrm{feats}}$ got connected through an undirected edge $e_{r,jk}\in\mathbb{E}_{r,\mathrm{feats}}$ with a weight $w_{r,jk}=w_{r,kj}\in\mathbb{R}_{\geq 0}$ reflecting their features differences. Accordingly, the degree of each sample (vertex) $v_{r,j}\in\mathbb{V}_{r,\mathrm{feats}}$ over the subgraph $\mathcal{G}_{r,\mathrm{feats}}$ was
\begin{equation}
\label{eq:degVerFeatGraph2}
d_{r,j}=\sum_{e_{r,jk}\in\mathbb{E}_{r,\mathrm{feats}}}w_{r,jk}.
\end{equation}

\begin{figure}[t!]
\begin{center}
\includegraphics[width=0.9\textwidth,height=18cm]{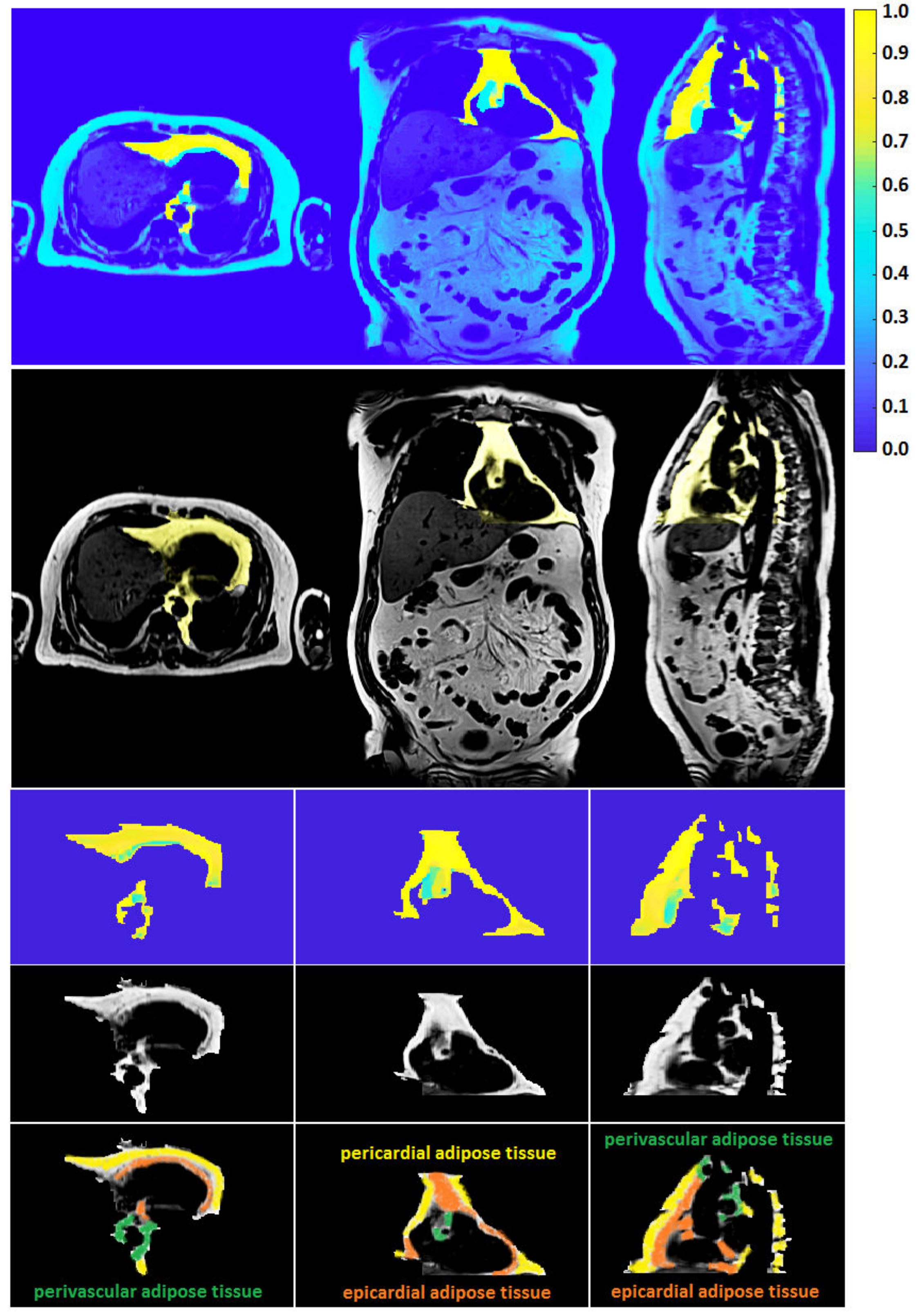}
\caption{Voxelwise classification probabilities estimated by the proposed forest for epicardial (orange), pericardial (yellow), and cardiac perivascular (green) adipose tissues and the resulting segmentations on various slices of a fat image.}
\label{fig:PriorProbCAT}
\end{center}
\end{figure}

The prior probabilities of the unlabeled samples $\mathbb{D}_{r,\mathrm{unl}}\subset\mathbb{D}_r$ formed a matrix $\mathbf{A}_{r,\mathrm{unl}}={[\mathbf{a}_{r,c,\mathrm{unl}}]}_{:,c}={[\mathbf{a}_{r,j,\mathrm{unl}}]}_{j,:}={[a_{r,j,c,\mathrm{unl}}]}_{j,c}$ of dimension $|\mathbb{D}_{r,\mathrm{unl}}|\times n_{\mathrm{clas}}$ and got encoded into the aspatial prior-based subgraph $\mathcal{G}_{r,\mathrm{prior}}=(\mathbb{V}_{r,\mathrm{prior}},\mathbb{E}_{r,\mathrm{prior}})$. To this end, $\mathbb{V}_{r,\mathrm{feats}}$ got partitioned into the labeled $\mathbb{V}_{r,\mathrm{lab}}$ and unlabeled $\mathbb{V}_{r,\mathrm{unl}}$ vertices with $\mathbb{V}_{r,\mathrm{lab}}\cup\mathbb{V}_{r,\mathrm{unl}}=\mathbb{V}_{r,\mathrm{feats}}$ and $\mathbb{V}_{r,\mathrm{lab}}\cap\mathbb{V}_{r,\mathrm{unl}}=\emptyset$. Then, every vertex $v_{r,c}\in\mathbb{V}_{r,\mathrm{prior}}$ represented a class $c\in\mathbb{L}$ and got connected to every vertex $v_{r,j}\in\mathbb{V}_{r,\mathrm{unl}}$ via an undirected edge $e_{r,jc}\in\mathbb{E}_{r,\mathrm{prior}}$ with a weight
\begin{equation}
\label{eq:weightPriors2}
w_{r,jc}=w_{r,cj}=\lambda_{r,\mathrm{prior}}\cdot a_{r,j,c}~~~~\text{with}~~~~|\mathbb{V}_{r,\mathrm{prior}}|=|\mathbb{L}|=n_{\mathrm{clas}}.
\end{equation}
Here, $\lambda_{r,\mathrm{prior}}\in\mathbb{R}_{+}$ was a resolution-specific hyperparameter.

\begin{figure}[t!]
\begin{center}
\includegraphics[width=0.9\textwidth]{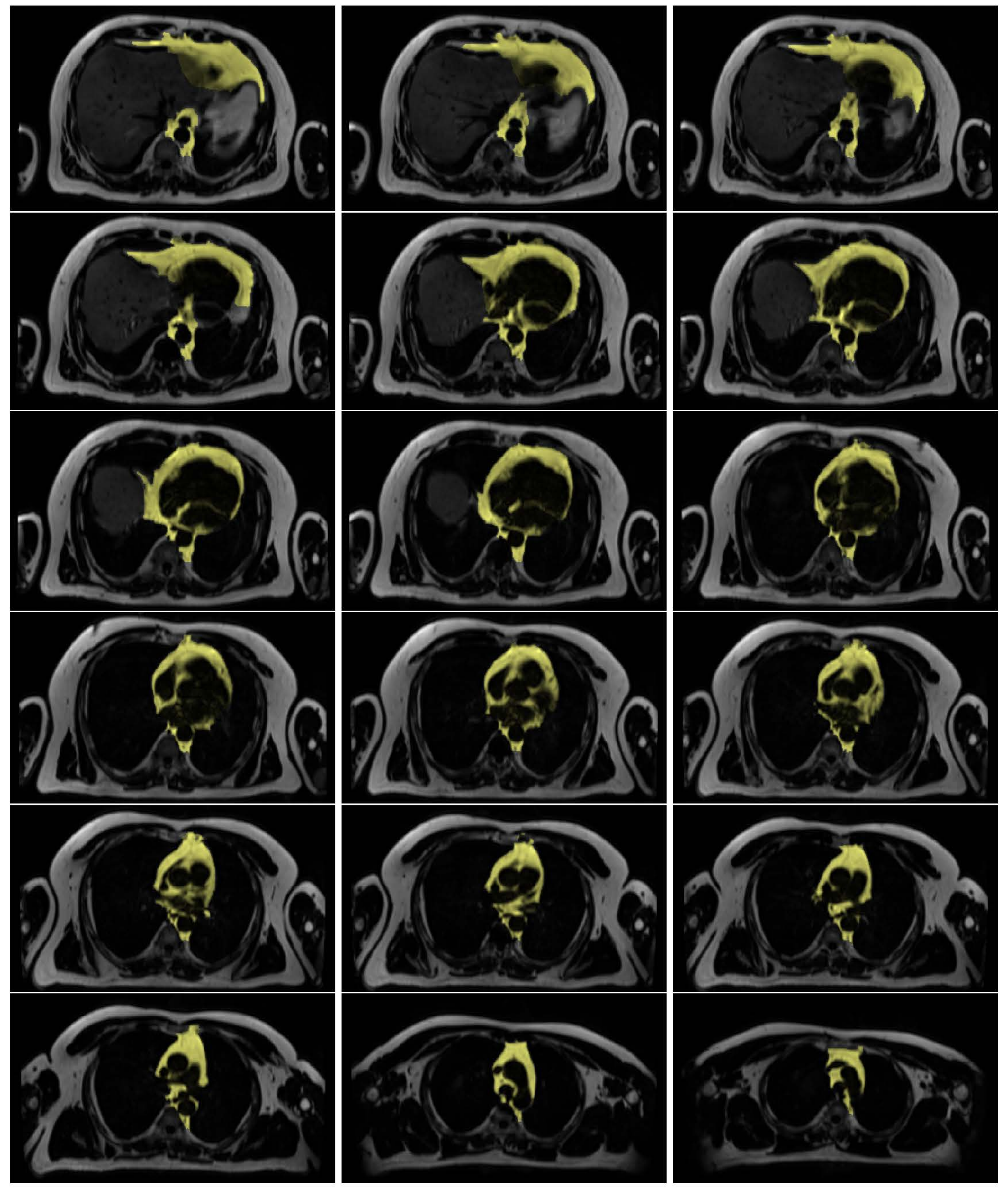}
\caption{Overall segmentations of the cardiac adipose tissues based on the voxelwise classification probabilities estimated by the proposed forest on some axial slices of a fat image.}
\label{fig:AxialSegCAT}
\end{center}
\end{figure}

The classification priors of the samples could be estimated by any method such as the multiatlas registration or the hierarchical random forest classifier proposed in \cite{Fallah2018a,Fallah2018p,Fallah2019a,FallahJ2019}. This happened when the classification probabilities estimated by those methods were not accurate enough to be considered as final classification probabilities (posteriors) rather as initial classification probabilities (priors). For example, the classification probabilities estimated by the proposed forest could not differentiate different kinds of cardiac adipose tissues on fat-water MR images. They could only differentiate the overall adipose tissues from nonadipose tissues. This was due to lack of spatial information in the forest, similar contrasts and features of the cardiac adipose tissues, and weak boundaries between them. The cardiac adipose tissues were spatially close to each other and separated only through thin septa which were hardly detectable under spatial resolution of standard clinical scanners at 3 T. \autoref{fig:PriorProbCAT} shows voxelwise classification probabilities estimated by the proposed forest for the cardiac adipose tissues and the resulting segmentations on various slices of a fat image. \autoref{fig:AxialSegCAT} shows overall segmentations of the cardiac adipose tissues based on the voxelwise classification probabilities estimated by the proposed forest on some axial slices of a fat image.
Despite inaccuracies of the priors, they, features, and the spatial (neighborhood) relationships of the samples could pave the way for another classifier such as the neighborhood graph $\mathcal{G}_r$ to classify the samples more accurately. For example, the priors could approximately localize an addressed object in an image and thereby speed up or reduce the complexity of its accurate segmentation. The more accurate classification probabilities were called the \textbf{posteriors}. Accordingly, the neighborhood graph $\mathcal{G}_r$ was supposed to estimate the classification posteriors $\mathbf{p}_{r,j}={[p_{r,j,c}]}_{c\in\mathbb{L}}$ of every sample (vertex) $v_{r,j}\in\mathbb{V}_{r,\mathrm{feats}}$ with regard to its features, its priors, and its spatial (neighborhood) relationships with other samples of the same resolution $r$.

Commonly used algorithms for solving classification problems on a CRF-based neighborhood graph $\mathcal{G}_r=\mathcal{G}_{r,\mathrm{feats}}\bigcup\mathcal{G}_{r,\mathrm{prior}}$ were the Graph-Cut \cite{Boykov2006} and the random walker algorithm \cite{Grady2008}. The Graph-Cut algorithm estimated binary classification labels of the unlabeled samples $\mathbb{D}_{r,\mathrm{unl}}\subset\mathbb{D}_r$ by minimizing the energy function (sum of potentials) of the CRF. This energy function was formed from the relationships of the prelabeled $\mathbb{D}_{r,\mathrm{lab}}\subset\mathbb{D}_r$ and unlabeled samples $\mathbb{D}_{r,\mathrm{unl}}\subset\mathbb{D}_r$ over the spatial feature-based subgraph $\mathcal{G}_{r,\mathrm{feats}}=(\mathbb{V}_{r,\mathrm{feats}},\mathbb{E}_{r,\mathrm{feats}})$. The minimization was done by using a primal-dual algorithm based on a max-flow min-cut optimization over a directed capacitated graph. This graph was built over the spatial feature-based subgraph $\mathcal{G}_{r,\mathrm{feats}}$ and included the vertices $\mathbb{V}_{r,\mathrm{feats}}$ plus a source and a target. The target represented the addressed foreground class $c\in\mathbb{L}$ and was connected to every sample (vertex) $v_{r,j}\in\mathbb{V}_{r,\mathrm{feats}}$ via an edge with a weight $\lambda_{r,\mathrm{prior}}\cdot a_{r,j,c}$.
The primal-dual algorithm based on a max-flow min-cut optimization had an efficient implementation proposed in \cite{Komodakis2008,Zhang2020}. However, the max-flow min-cut optimization for a multiclass classification was NP-hard. Thus, to solve a multiclass classification problem with it, this problem should be decomposed into a series of binary (one-vs-all) classifications. Then, for each classification, a directed capacitated graph should be built and a max-flow min-cut optimization should be performed. In addition, the minimization of the energy function of the CRF returned discrete-valued labels but no classification probabilities and thus no measure about the classification confidence. Moreover, the min-cut could only minimally separate different classes in spatial domain. Thus, it could fail if the number and distribution of the prelabeled samples (seeds) were not enough or the classification priors were not helpful. Furthermore, multiple min-cuts could emerge and there was no guarantee to find the optimum one. This increased the sensitivity of the classifier to noise and perturbations.

The random walker algorithm, proposed in \cite{Grady2008}, mitigated the above shortcomings by casting the classification problem into a combinatorial Dirichlet problem whose boundary conditions were the prelabeled samples (seeds) and/or the classification priors of the unlabeled samples. This formulation relaxed the integer (binary) constraints of the energy function of the CRF and allowed to simultaneously compute the classification probabilities (posteriors) of the unlabeled samples $\mathbb{D}_{r,\mathrm{unl}}$ with regard to $n_{\mathrm{clas}}$ classes. These posteriors formed a matrix $\hat{\mathbf{P}}_{r,\mathrm{unl}}={[\hat{\mathbf{p}}_{r,c,\mathrm{unl}}]}_{:,c}={[\hat{\mathbf{p}}_{r,j,\mathrm{unl}}]}_{j,:}={[\hat{p}_{r,j,c,\mathrm{unl}}]}_{j,c}$ of dimension $|\mathbb{D}_{r,\mathrm{unl}}|\times n_{\mathrm{clas}}$.
The posteriors of the prelabeled samples $\mathbb{D}_{r,\mathrm{lab}}$ were known and were simply the one-hot-encoding of their reference (ground truth) labels. These posteriors formed a $|\mathbb{D}_{r,\mathrm{lab}}|\times n_{\mathrm{clas}}$ matrix $\hat{\mathbf{P}}_{r,\mathrm{lab}}={[\hat{\mathbf{p}}_{r,c,\mathrm{lab}}]}_{:,c}={[\hat{\mathbf{p}}_{r,j,\mathrm{lab}}]}_{j,:}={[\hat{p}_{r,j,c,\mathrm{lab}}]}_{j,c}$ with
\begin{equation}
\label{eq:postLabSmpls2}
\hat{p}_{r,j,c,\mathrm{lab}}=\begin{cases}1-\epsilon&~\text{if}~c=l_{r,j}\\\epsilon/(n_{\mathrm{clas}}-1)&~\text{if}~c\neq l_{r,j}\end{cases}~~~~~~~~~\epsilon=0.0001
\end{equation}
and $l_{r,j}\in\mathbb{L}$ being the reference label of the sample (vertex) $v_{r,j}\in\mathbb{V}_{r,\mathrm{lab}}\subset\mathbb{V}_{r,\mathrm{feats}}$.

The overall posteriors formed a matrix $\hat{\mathbf{P}}_r=\begin{bmatrix}\hat{\mathbf{P}}_{r,\mathrm{lab}}\\\hat{\mathbf{P}}_{r,\mathrm{unl}}\end{bmatrix}={[\hat{\mathbf{p}}_{r,c}]}_{:,c}={[\hat{\mathbf{p}}_{r,j}]}_{j,:}={[\hat{p}_{r,j,c}]}_{j,c}$ of dimension $|\mathbb{D}_r|\times n_{\mathrm{clas}}=|\mathbb{V}_{r,\mathrm{feats}}|\times n_{\mathrm{clas}}$ and were the domain of the discrete (combinatorial) Dirichlet integral of the neighborhood (random walker) graph $\mathcal{G}_r=\mathcal{G}_{r,\mathrm{feats}}\bigcup\mathcal{G}_{r,\mathrm{prior}}$.

The combinatorial Dirichlet integral of the graph $\mathcal{G}_r$ was given by
\begin{subequations}
\label{eq:CombDirichInt2}
\begin{align*}
\mathcal{L}_{\mathrm{Dirich}}(\hat{\mathbf{P}}_r)&=\frac{1}{2}\sum_{c=1}^{n_{\mathrm{clas}}-1}\Big[\sum_{\substack{e_{r,jk}\in\\\mathbb{E}_{r,\mathrm{feats}}}}w_{r,jk}\cdot(\hat{p}_{r,j,c}-\hat{p}_{r,k,c})^2+\sum_{\substack{v_{r,j}\in\\\mathbb{V}_{r,\mathrm{unl}}}}\lambda_{r,\mathrm{prior}}\cdot(\hat{p}_{r,j,c}-a_{r,j,c})^2\Big]\\
&=\frac{1}{2}\underbrace{\begin{bmatrix}\hat{\mathbf{P}}_{r,\mathrm{lab}}^T&\hat{\mathbf{P}}_{r,\mathrm{unl}}^T\end{bmatrix}}_{\hat{\mathbf{P}}_r^T}\times\underbrace{\begin{bmatrix}\boldsymbol{\Gamma}_{r,\mathrm{lab}}&\boldsymbol{\Gamma}_{r,\mathrm{ula}}\\\boldsymbol{\Gamma}_{r,\mathrm{ula}}^T&\boldsymbol{\Gamma}_{r,\mathrm{unl}}\end{bmatrix}}_{\boldsymbol{\Gamma}_r}\times\underbrace{\begin{bmatrix}\hat{\mathbf{P}}_{r,\mathrm{lab}}\\\hat{\mathbf{P}}_{r,\mathrm{unl}}\end{bmatrix}}_{\hat{\mathbf{P}}_r}\tag{\ref{eq:CombDirichInt2}}
\end{align*}
\end{subequations}
with $\hat{p}_{r,j,c}\in(0,1)$ being the posterior of the sample (vertex) $v_{r,j}\in\mathbb{V}_{r,\mathrm{feats}}$ with regard to the class $c\in\mathbb{L}$, $w_{r,jk}=w_{r,kj}\in\mathbb{R}_{\geq 0}$ being the weight of the edge $e_{r,jk}\in\mathbb{E}_{r,\mathrm{feats}}$, and $\boldsymbol{\Gamma}_r$ being the $|\mathbb{D}_r|\times|\mathbb{D}_r|$ combinatorial Laplace-Beltrami matrix of the graph $\mathcal{G}_r$. This matrix was sparse symmetric positive semidefinite and was given by
\begin{equation}
\label{eq:combLapMtrx2}
\gamma_{r,j,k}=\begin{cases}
d_{r,j}+\lambda_{r,\mathrm{prior}}&\text{if}~v_{r,j}=v_{r,k}\\
-w_{r,jk}&\text{if}~e_{r,jk}\in\mathbb{E}_{r,\mathrm{feats}}\\
0&\text{otherwise}
\end{cases}~~~~~\boldsymbol{\Gamma}_r={[\gamma_{r,j,k}]}_{j,k}
\end{equation}
with $d_{r,j}=\sum_{e_{r,jk}\in\mathbb{E}_{r,\mathrm{feats}}}w_{r,jk}$ being the degree of $v_{r,j}\in\mathbb{V}_{r,\mathrm{feats}}$ over the subgraph $\mathcal{G}_{r,\mathrm{feats}}$.

The integral (sum) in \eqref{eq:CombDirichInt2} ran over $(n_{\mathrm{clas}}-1)$ classes because the posteriors of each sample (vertex) were supposed to sum to one. Thus, the remaining posterior of each sample was one minus sum of the $(n_{\mathrm{clas}}-1)$ computed posteriors. The optimum classification posteriors $\mathbf{P}_r={[\mathbf{p}_{r,c}]}_{:,c}={[\mathbf{p}_{r,j}]}_{j,:}={[p_{r,j,c}]}_{j,c}$ were the minimizers of \eqref{eq:CombDirichInt2}. That is,
\begin{equation}
\label{eq:DirichMin2}
\mathbf{P}_r=\begin{bmatrix}\mathbf{P}_{r,\mathrm{lab}}\\\mathbf{P}_{r,\mathrm{unl}}\end{bmatrix}=\argmin_{\hat{\mathbf{P}}_r}~\mathcal{L}_{\mathrm{Dirich}}(\hat{\mathbf{P}}_r).
\end{equation}

By vanishing the derivative of \eqref{eq:CombDirichInt2} with respect to each $\hat{p}_{r,j,c}\in(0,1)$, one obtained
\begin{equation}
\label{eq:probEq2}
\hat{p}_{r,j,c}=\frac{\lambda_{r,\mathrm{prior}}\cdot a_{r,j,c}+\sum_{e_{r,jk}\in\mathbb{E}_{r,\mathrm{feats}}}w_{r,jk}\cdot\hat{p}_{r,k,c}}{d_{r,j}+\lambda_{r,\mathrm{prior}}}.
\end{equation}
Also, by vanishing the derivative of \eqref{eq:CombDirichInt2} with respect to $\hat{\mathbf{P}}_{r,\mathrm{unl}}$, one obtained
\begin{equation}
\label{eq:linEqsRandWalk2}
\begin{split}
\text{for~all~classes}:&~~~\boldsymbol{\Gamma}_{r,\mathrm{unl}}\times\mathbf{P}_{r,\mathrm{unl}}=-\boldsymbol{\Gamma}_{r,\mathrm{ula}}^T\times\mathbf{P}_{r,\mathrm{lab}}+\lambda_{r,\mathrm{prior}}\times\mathbf{A}_{r,\mathrm{unl}}\\
\text{for~one~class}~c\in\mathbb{L}:&~~~\boldsymbol{\Gamma}_{r,\mathrm{unl}}\times\mathbf{p}_{r,c,\mathrm{unl}}=-\boldsymbol{\Gamma}_{r,\mathrm{ula}}^T\times\mathbf{p}_{r,c,\mathrm{lab}}+\lambda_{r,\mathrm{prior}}\times\mathbf{a}_{r,c,\mathrm{unl}}.
\end{split}
\end{equation}

Accordingly, the random walker algorithm formulated a probabilistic multiclass classification as the closed form analytical solution of a system of sparse linear equations given by \eqref{eq:linEqsRandWalk2}. The sparsity of this system stemmed from the sparsity of the matrix $\boldsymbol{\Gamma}_r$ whose elements were the coefficients of the system. The known information of this system were the prelabeled samples (seeds) and/or the classification priors of the unlabeled samples. This way, the random walker algorithm enabled a more accurate but less complex multiclass classification without encountering with the NP-hardness of the max-flow min-cut optimization. It could also avoid the trivial solutions, the uncertainties, and the sensitivities of the min-cut to the number and distribution of the prelabeled samples or the helpfulness of the priors. These resulted in a less sensitivity to noise and perturbations and a better delineation of weak boundaries between different classes (e.g. objects) in spatial domain (e.g. image).

In the original random walker algorithm for image segmentation, reference (ground truth) labels of the prelabeled validation/test samples $\mathbb{D}_{r,\mathrm{lab}}\subset\mathbb{D}_r$ should be manually and interactively provided by a user of the algorithm \cite{Grady2008}. These annotations were tedious and cumbersome especially when a large set of images should be segmented. Hence, unless otherwise specified, we assumed that all the samples $\mathbb{D}_r$ were unlabeled and thus accompanied with their prior classification probabilities. This implied $\mathbb{D}_{r,\mathrm{lab}}=\emptyset$, $\mathbb{D}_{r,\mathrm{unl}}=\mathbb{D}_r$, $\mathbb{V}_{r,\mathrm{lab}}=\emptyset$, $\mathbb{V}_{r,\mathrm{unl}}=\mathbb{V}_{r,\mathrm{feats}}$, $\mathbf{P}_{r,\mathrm{unl}}=\mathbf{P}_r$, $\mathbf{A}_{r,\mathrm{unl}}=\mathbf{A}_r={[\mathbf{a}_{r,c}]}_{:,c}={[\mathbf{a}_{r,j}]}_{j,:}={[a_{r,j,c}]}_{j,c}$, $\boldsymbol{\Gamma}_{r,\mathrm{ula}}=\boldsymbol{\Gamma}_{r,\mathrm{lab}}=\emptyset$, and $\boldsymbol{\Gamma}_{r,\mathrm{unl}}=\boldsymbol{\Gamma}_r$. In this case, \eqref{eq:linEqsRandWalk2} converted to
\begin{equation}
\label{eq:linEqsUnlab2}
\begin{split}
\text{for~all~classes}:&~~~\boldsymbol{\Gamma}_r\times\mathbf{P}_r=\lambda_{r,\mathrm{prior}}\times\mathbf{A}_r\\
\text{for~one~class}~c\in\mathbb{L}:&~~~\boldsymbol{\Gamma}_r\times\mathbf{p}_{r,c}=\lambda_{r,\mathrm{prior}}\times\mathbf{a}_{r,c}.
\end{split}
\end{equation}

\section{Outline of Contributions}
\label{sec:OutlineGraphs}
To estimate the classification posteriors $\mathbf{P}_r={[\mathbf{p}_{r,c}]}_{:,c}={[\mathbf{p}_{r,j}]}_{j,:}={[p_{r,j,c}]}_{j,c}$ of unlabeled validation/test samples $\mathbb{D}_r=\mathbb{D}_{r,\mathrm{val}}\cup\mathbb{D}_{r,\mathrm{test}}$ of each resolution $r\in\{0,\cdots,n_{\mathrm{lay}}\}$ from their features, priors, and spatial (neighborhood) relationships, we proposed three variants of the neighborhood graph $\mathcal{G}_r=\mathcal{G}_{r,\mathrm{feats}}\bigcup\mathcal{G}_{r,\mathrm{prior}}$. These graphs were referred to as
\begin{itemize}[leftmargin=*]
\item\textbf{feature- and prior-based graph}
\item\textbf{constrained feature- and prior-based graph}
\item\textbf{guided feature- and prior-based graph}.
\end{itemize}
They expanded on the random walker graph for image segemenetaion proposed in \cite{Grady2008} by using different mechanisms to derive the edge weights ${\{w_{r,jk}=w_{r,kj}\in\mathbb{R}_{\geq 0}\}}_{e_{r,jk}\in\mathbb{E}_{r,\mathrm{feats}}}$ of their spatial feature-based subgraph $\mathcal{G}_{r,\mathrm{feats}}=(\mathbb{V}_{r,\mathrm{feats}},\mathbb{E}_{r,\mathrm{feats}})$. In the original random walker graph, feature of every sample (vertex) $v_{r,j}\in\mathbb{V}_{r,\mathrm{feats}}$ was the average intensity\footnote{The original random walker algorithm, proposed in \cite{Grady2008}, was for single-channel data sets (images).} $\iota_{r,j}$ of its patch. Then, the weight $w_{r,jk}=w_{r,kj}\in\mathbb{R}_{\geq 0}$ of every edge $e_{r,jk}\in\mathbb{E}_{r,\mathrm{feats}}$ was a radial basis function of the differences of the average intensities of the spatially neighboring samples $v_{r,j},~v_{r,k}\in\mathbb{V}_{r,\mathrm{feats}}$ as
\begin{equation}
\label{eq:weightFeats2}
w_{r,jk}=w_{r,kj}=\mathrm{exp}\big(-(\iota_{r,j}-\iota_{r,k})^2\big).
\end{equation}
This way, each weight $w_{r,jk}=w_{r,kj}\in\mathbb{R}_{\geq 0}$ represented the probability that a random walker would pass through the edge $e_{r,jk}\in\mathbb{E}_{r,\mathrm{feats}}$. Thus, it should help the random walker to avoid crossing (weak) boundaries between different classes in the spatial domain.

To enhance detection of weak boundaries between different classes, we defined every edge weight $w_{r,jk}=w_{r,kj}\in\mathbb{R}_{\geq 0}$ based on the Tukey's function of the features differences of the connected samples multiplied by the classification reliabilities of the samples. These edge weights underpinned an \textbf{implicit boundary detection capability}.

In regression problems, the Tukey's function or its derivative were used as the objective function of neural networks trained (optimized) with gradient descent and backpropagation. This was due to the outlier detection/removal capability of those functions which in turn prevented to bias the training (optimization) with the large gradients caused by the outliers. This capability was controlled by a tuning parameter \cite{Beaton1974,Belagiannis2015}. In our case, the outliers were the boundaries between different classes in spatial domain.
According to robust statistics \cite{Black1998,Rousseeuw1987}, we derived the tuning parameter of the Tukey's function from the median of absolute deviations of the features differences from the median of the features differences of the connected samples.

\begin{figure}[t!]
\begin{center}
\includegraphics[width=1.0\textwidth]{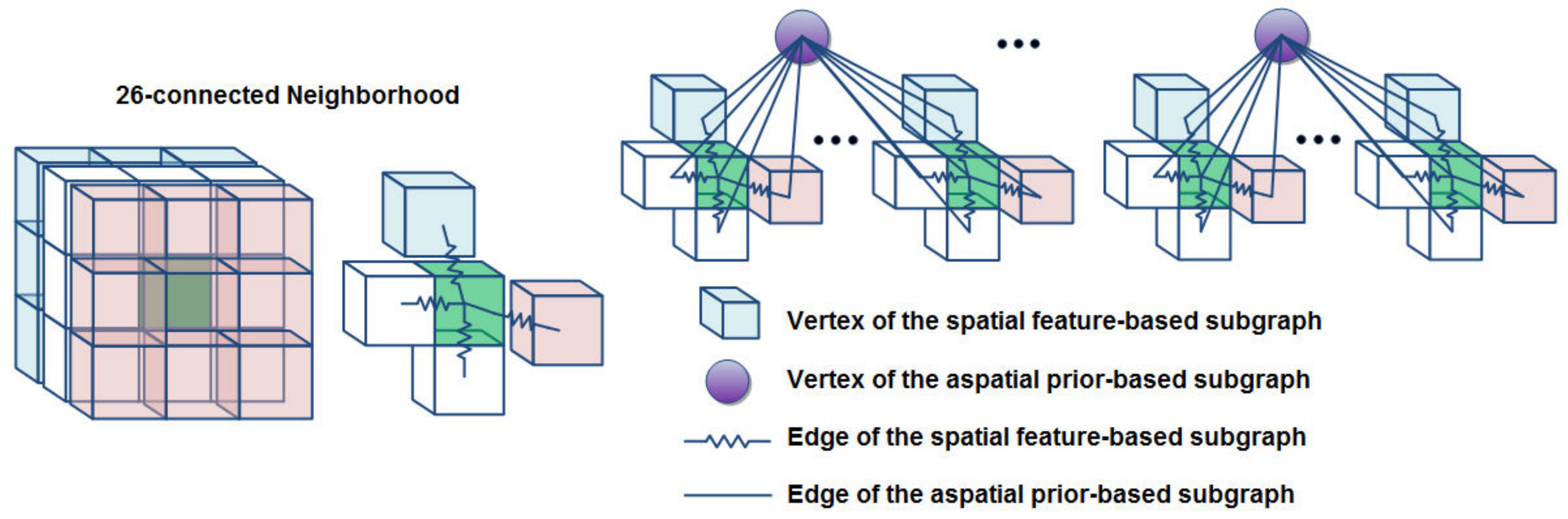}
\caption{The 26-connected neighborhood and a portion of the graph $\mathcal{G}_r$ involving this kind of neighborhood in its spatial feature-based subgraph $\mathcal{G}_{r,\mathrm{feats}}$.}
\label{fig:NeighGraph}
\end{center}
\end{figure}

Furthermore, in the original random walker graph for image segmentation, every sample had 8 direct spatial neighbors which was called a second-order or an 8-connected neighborhood in 2D. We extended this to a second-order or a 26-connected neighborhood in 3D which resulted in $26$ direct spatial neighbors for each sample (vertex) $v_{r,j}\in\mathbb{V}_{r,\mathrm{feats}}$ on the subgraph $\mathcal{G}_{r,\mathrm{feats}}$. \autoref{fig:NeighGraph} shows the 26-connected neighborhood and a portion of the graph $\mathcal{G}_r$ involving this kind of neighborhood in its spatial feature-based subgraph $\mathcal{G}_{r,\mathrm{feats}}$.
The 26-connected neighborhood allowed the information to propagate across the vertices (samples) of $\mathcal{G}_{r,\mathrm{feats}}$ without imposing the complexity of nearest neighbor calculation or any other clustering (grouping) method. In addition, it kept the topology of the subgraph $\mathcal{G}_{r,\mathrm{feats}}$ regular and homogeneous to the extent that the derivation of the real valued edge weights of $\mathcal{G}_{r,\mathrm{feats}}$ did not result in vanishing of some weights. That is, it could happen that two samples were 26-connected neighbors of each other but their features were so different that their connecting edge weight vanished. In this case, the samples were disconnected.

The constrained feature- and prior-based graph imposed additional constraints to the edge weights of the spatial feature-based subgraph to enhance the classification performance while reducing its complexity by reducing the number of unknown classifications. These constraints were derived from the classification priors of the samples and/or the \textbf{boundaries explicitly detected} between different classes in spatial domain. The explicit boundary detection was through an application of the 3D Sobel operator to each intensity channel of the samples.

The guided feature- and prior-based graph incorporated a diffusion-based susceptible-infected-recovered (SIR) model into the edge weights of the spatial feature-based subgraph. This model allowed to use additional application-specific information for guiding the classifications and thereby tackling classification challenges \cite{Bampis2017}. We derived it for the challenging task of segmenting cardiac adipose tissues on volumetric fat-water MR images. For this, the additional information got inferred from the orientations of the mode of the histogram of oriented gradients (HOG) of the fat channels of the samples and the surface curvatures of the cardiac structure segmented over the corresponding volumetric water image.
This way, each neighborhood graph $\mathcal{G}_r$ estimated the classification posteriors $\mathbf{P}_r$ of the samples $\mathbb{D}_r$ of the resolution $r\in\{0,\cdots,n_{\mathrm{lay}}\}$. To this end, all of these samples should have a \textbf{common set of feature types specific to the resolution $r$}.

\begin{figure}[t!]
\begin{center}
\includegraphics[width=1.0\textwidth,height=18cm]{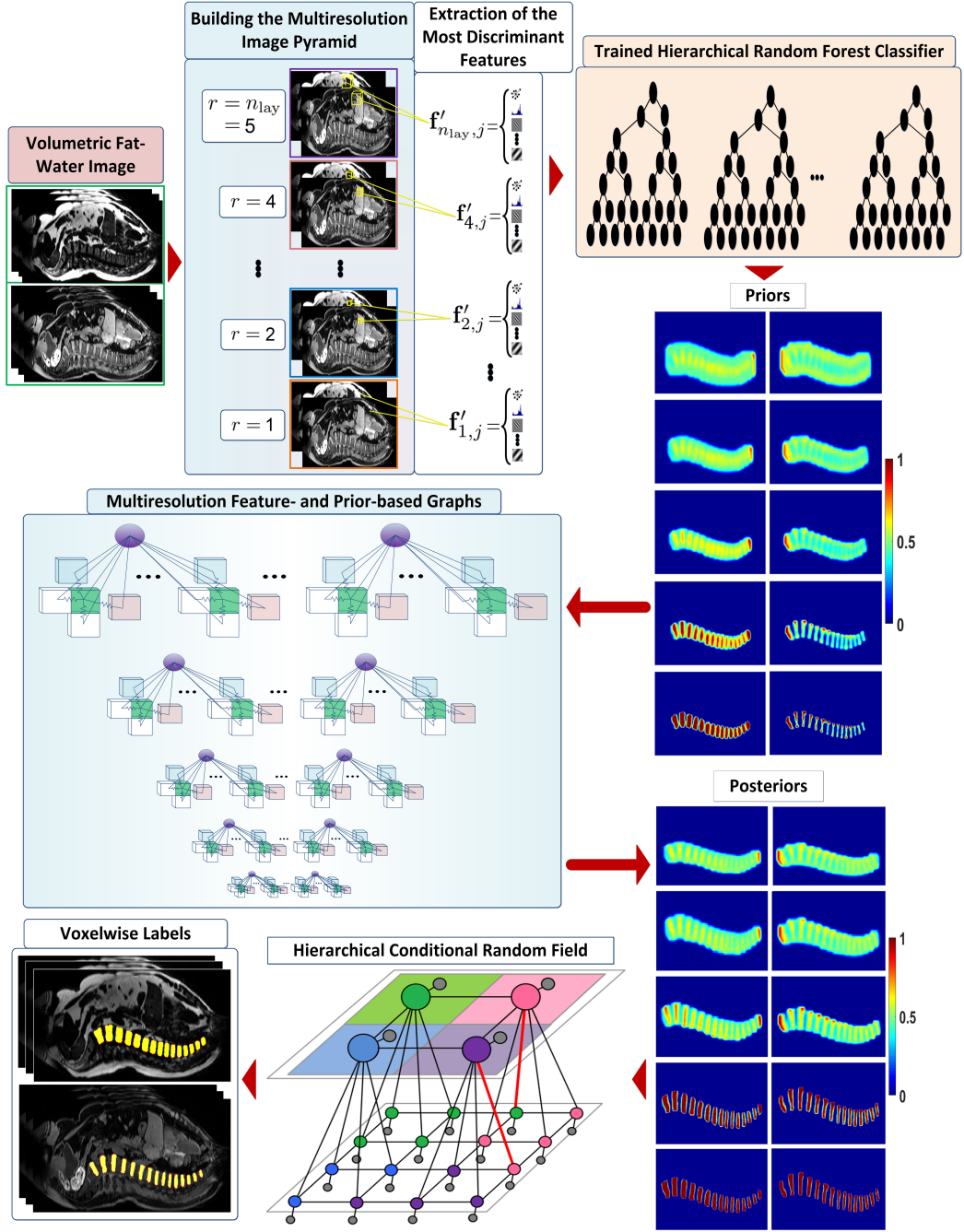}
\caption{The proposed classification pipeline for segmenting a volumetric fat-water image.}
\label{fig:ClassMultResGraph}
\end{center}
\end{figure}

To estimate the posteriors ${\{\mathbf{P}_r\}}_{r=0}^{n_{\mathrm{lay}}}$ of the multiresolution samples ${\{\mathbb{D}_r\}}_{r=0}^{n_{\mathrm{lay}}}$, a stack of multiresolution graphs ${\{\mathcal{G}_r\}}_{r=0}^{n_{\mathrm{lay}}}$ was needed. These graphs were disconnected from each other and thus acted independently. Hence, they could not incorporate the inter-resolution hierarchical relationships of the multiresolution samples into their classifications. To incorporate these relationships into the classifications of the multiresolution samples ${\{\mathbb{D}_r\}}_{r=0}^{n_{\mathrm{lay}}}$, we fused the multiresolution posteriors ${\{\mathbf{P}_r\}}_{r=0}^{n_{\mathrm{lay}}}$ into the hierarchically consistent labels ${\big\{\mathbf{l}^{*}_r={[l^{*}_{r,j}\in\mathbb{L}]}_j\big\}}_{r=0}^{n_{\mathrm{lay}}}$ by using these relationships. This was done over a weighted undirected graph $\mathcal{G}_{\mathrm{hcrf}}=(\mathbb{V}_{\mathrm{hcrf}},\mathbb{E}_{\mathrm{hcrf}})$ of a hierarchical conditional random field (HCRF). The graph represented the multiresolution samples by its vertices $\mathbb{V}_{\mathrm{hcrf}}$ and encoded the inter-resolution hierarchical relationships of these samples by the weights of its edges $\mathbb{E}_{\mathrm{hcrf}}$. This way, it connected the graphs ${\{\mathcal{G}_r\}}_{r=0}^{n_{\mathrm{lay}}}$ with regard to those relationships. To enable this connection, the samples ${\{\mathbb{D}_r\}}_{r=0}^{n_{\mathrm{lay}}}$ should have a common set of \textbf{resolution-independent feature types}. Being undirected allowed the graph $\mathcal{G}_{\mathrm{hcrf}}$ to encode the hierarchical relationships in both coarse-to-fine (parent-to-child) and fine-to-coarse (child-to-parent) directions. Similar to our neighborhood graphs, we defined the edge weights of the graph $\mathcal{G}_{\mathrm{hcrf}}$ based on the Tukey's function of the features differences of the connected samples multiplied by their classification reliabilities.

The proposed graphs used the information contained in the multiresolution validation/test samples ${\{\mathbb{D}_r=\mathbb{D}_{r,\mathrm{val}}\cup\mathbb{D}_{r,\mathrm{test}}\}}_{r=0}^{n_{\mathrm{lay}}}$. We assumed that these samples were already processed by a hierarchical random forest classifier such as the forest proposed in \cite{Fallah2018a,Fallah2018p,Fallah2019a,FallahJ2019}. At the end of this process, each validation/test sample $v_{r,j}\in\mathbb{D}_r$ had its
\begin{enumerate}[label={(\arabic*)},leftmargin=*]
\item\label{extSmpls}fat-water patch $\rho_{r,j}$ of resolution $r\in\{0,\cdots,n_{\mathrm{lay}}\}$
\item vector of normalized prior probabilities $\mathbf{a}_{r,j}={\big[a_{r,j,c}\in(0,1)\big]}_{c\in\mathbb{L}}$ resulted from the classification probabilities estimated over the forest
\item indicator of classification reliability $h^{*}_{r,j}\in(0,1]$ given by \eqref{eq:diffRandForest}
\item vector of consolidated resolution-specific features $\tilde{\mathbf{f}}_{r,j}$
\item vector of consolidated resolution-independent features $\hat{\mathbf{f}}_{r,j}$
\item 26 spatial neighbors
\item hierarchical parent in the $(r+1)^{\mathrm{th}}$ resolution layer
\item hierarchical children in the $(r-1)^{\mathrm{th}}$ resolution layer.
\end{enumerate}
More specifically, the multiresolution neighborhood graphs ${\{\mathcal{G}_r\}}_{r=0}^{n_{\mathrm{lay}}}$ estimated the classification posteriors ${\{\mathbf{P}_r\}}_{r=0}^{n_{\mathrm{lay}}}$ of the multiresolution samples ${\{\mathbb{D}_r\}}_{r=0}^{n_{\mathrm{lay}}}$ by using the aforementioned information except for (5), (7), and (8). The hierarchical graph $\mathcal{G}_{\mathrm{hcrf}}$ fused these multiresolution posteriors into the hierarchically consistent labels by using the information indicated by (5), (7), and (8). This way, the proposed graphs could improve the classifications of the hierarchical quadratic random forest classifier proposed in \cite{Fallah2018a,Fallah2018p,Fallah2019a,FallahJ2019}. This resulted in a classification pipeline depicted in \autoref{fig:ClassMultResGraph}. At the end of this pipeline, the hierarchically consistent labels ${\big\{\mathbf{l}^{*}_r={[l^{*}_{r,j}\in\mathbb{L}]}_j\big\}}_{r=0}^{n_{\mathrm{lay}}}$ got evaluated against their corresponding reference labels ${\big\{\mathbf{l}_r={[l_{r,j}\in\mathbb{L}]}_j\big\}}_{r=0}^{n_{\mathrm{lay}}}$. If $\mathbb{D}_r=\mathbb{D}_{r,\mathrm{val}}$, then these evaluations were with regard to the selected hyperparameter values for the graphs. If $\mathbb{D}_r=\mathbb{D}_{r,\mathrm{test}}$, then these evaluations revealed the overall classification performance of the proposed pipeline after optimizing all the parameters.

\section{Feature- and Prior-based Graph}
\label{sec:FeatPriorGraph}
\begin{figure}[t!]
\begin{center}
\includegraphics[width=0.8\textwidth]{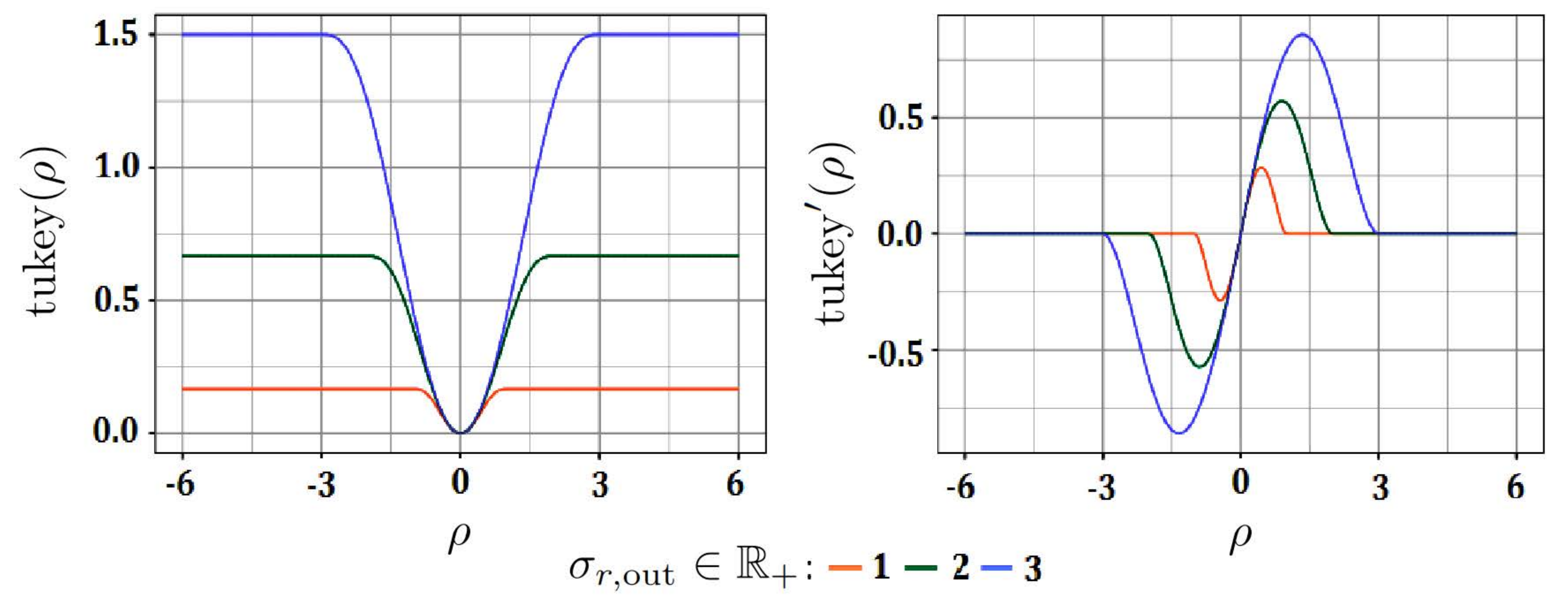}
\caption{The Tukey's function of a residual $\rho\in\mathbb{R}$ (left side) and its derivative (right side) for various values of the tuning parameter $\sigma_{r,\mathrm{out}}\in\mathbb{R}_{+}$.}
\label{fig:TukeyDerivFunc}
\end{center}
\end{figure}

A direct extension of \eqref{eq:weightFeats2} was to use the vector of resolution-specific features $\tilde{\mathbf{f}}_{r,j}$ of every sample (vertex) $v_{r,j}\in\mathbb{V}_{r,\mathrm{feats}}$ instead of its intensity $\iota_{r,j}$. This gave
\begin{equation}
\label{eq:weightFeats3}
w_{r,jk}=w_{r,kj}=\mathrm{exp}\Big(-{\|\tilde{\mathbf{f}}_{r,j}-\tilde{\mathbf{f}}_{r,k}\|}_2^2\Big)
\end{equation}
with ${\|\cdot\|}_2$ being the $l_2$ norm. We further extended \eqref{eq:weightFeats3} by using the Tukey's function of the features differences of the 26-connected samples multiplied by the classification reliabilities of the samples. This extension aimed to enhance detection of weak boundaries between different classes in spatial domain and relied on the assumption that these boundaries formed \textbf{outliers}. Thus, the better the outlier detection was, the better the boundary detection would be. Accordingly, we used the outlier detection/removal capability of the Tukey's function. The Tukey's function of a residual $\rho\in\mathbb{R}$ and its derivative were given by
\begin{equation}
\label{eq:TukeyFunc}
\mathrm{tukey}(\rho)=\begin{cases}\frac{\sigma_{r,\mathrm{out}}^2}{6}\cdot\Big(1-\big[1-(\frac{\rho}{\sigma_{r,\mathrm{out}}})^2\big]^3\Big)&\text{if}~|\rho|\leq\sigma_{r,\mathrm{out}}\\
\frac{\sigma_{r,\mathrm{out}}^2}{6}&\text{otherwise}
\end{cases}
\end{equation}
\begin{equation}
\label{eq:TukeyDerivFunc}
\mathrm{tukey}'(\rho)=\begin{cases}\rho\cdot\Big[1-(\frac{\rho}{\sigma_{r,\mathrm{out}}})^2\Big]^2&\text{if}~|\rho|\leq\sigma_{r,\mathrm{out}}\\
0&\text{otherwise}
\end{cases}
\end{equation}
with $\sigma_{r,\mathrm{out}}\in\mathbb{R}_{+}$ being a tuning parameter controlling the outlier detection/removal of these functions \cite{Beaton1974}. As shown in \autoref{fig:TukeyDerivFunc}, the parameter $\sigma_{r,\mathrm{out}}\in\mathbb{R}_{+}$ defined the range of inputs over which the Tukey's function/derivative got saturated/vanished. This way, the outliers manifested by large residuals or gradients could be suppressed. The minimum and the maximum of $\mathrm{tukey}'(\rho)$ occurred at $\rho=-\sigma_{r,\mathrm{out}}/\sqrt{5}$ and $\rho=\sigma_{r,\mathrm{out}}/\sqrt{5}$, respectively.

Since the neighborhood graph $\mathcal{G}_r$ was undirected, we only considered nonnegative residuals as the domain of the Tukey's function and derived each weight $w_{r,jk}=w_{r,kj}\in\mathbb{R}_{\geq 0}$ as
\begin{align}
\label{eq:weightTukeyFeats}
\begin{split}
w_{r,jk}=w_{r,kj}&=h^{*}_{r,j}~\cdot~h^{*}_{r,k}~\cdot~t_{r,jk}\\
t_{r,jk}=t_{r,kj}&=\begin{cases}\mathrm{exp}\Big(-\mathrm{tukey}\big({\|\tilde{\mathbf{f}}_{r,j}-\tilde{\mathbf{f}}_{r,k}\|}_1\big)\Big)&\text{if}~v_{r,k}\in\mathbb{N}_{r,j}\iff v_{r,j}\in\mathbb{N}_{r,k}\\0&\text{otherwise}
\end{cases}
\end{split}
\end{align}
with ${\|\cdot\|}_1$ being the $l_1$ norm, $\mathbb{N}_{r,j}$ being the set of samples (vertices) in the 26-connected neighborhood of $v_{r,j}\in\mathbb{D}_r$, and $h^{*}_{r,j}\in(0,1]$ being the indicator of classification reliability given by
\begin{subequations}
\label{eq:diffRandForest}
\begin{equation}
h^{*}_{r,j}=\mathrm{exp}(-\overline{h}_{r,j})~~~~~~~~~~~~\overline{h}_{r,j}=\frac{1}{n_{\mathrm{weak}}}\sum_{w=1}^{n_{\mathrm{weak}}}h_{r,w}
\end{equation}
\begin{equation}
h_{r,w}=h(\mathbb{D}_{r,w})=-\sum_{c\in\mathbb{L}}p_{r,w,c}\cdot\mathrm{log}_2(p_{r,w,c}):\text{empirical entropy}
\end{equation}
\begin{equation}
p_{r,w,c}=\frac{\mathrm{card}\big(\{v_{r,j}\in\mathbb{D}_{r,w}|l_{r,j}=c\}\big)}{\mathrm{card}(\mathbb{D}_{r,w})}:\text{empirical probability}
\end{equation}
\end{subequations}
for the sample (vertex) $v_{r,j}\in\mathbb{V}_{r,\mathrm{feats}}$. Here, $\mathbb{D}_{r,w}$ was the training samples of resolution $r$ used to optimize the decision nodes of the $r^{\mathrm{th}}$ resolution layer of the $w^{\mathrm{th}}$ weak classifier of the hierarchical random forest classifier. The multiplication $(h^{*}_{r,j}\cdot h^{*}_{r,k}\cdot t_{r,jk})$ stemmed from the fact that each validation/test sample $v_{r,j}\in\mathbb{D}_r$ had a fat-water patch designating its spatial region. In this region, a composition of different classes might exist and this composition was reflected by the voxelwise label heterogeneity of the patch. The higher this heterogeneity was, the more difficult/unreliable the classification of the sample would be. For the validation/test samples these label heterogeneities were not available at the time of their classification. The trained (optimized) forest inferred the classification reliabilities of these samples from the voxelwise label heterogeneities of the training patches previously processed by it.


If the residuals had a standard normal distribution (with zero mean and unit variance), then $\sigma_{r,\mathrm{out}}=4.6851$ would result in 95\% asymptotic efficiency for a least squares estimator\footnote{The asymptotic efficiency of an unbiased estimator was the percentage of the theoretical precision (concentration or inverse of variance) that the estimator could achieve by tending its number of samples to infinity. The asymptotic efficiency depended on the estimator and the distribution of the samples.} \cite{Belagiannis2015}. If the variance was not unit and the mean (average) was a reliable representative of the middle of the samples' distribution, then the unit variance got approximately fulfilled by dividing the residuals with the maximizer of the derivative of the Tukey's function (i.e. $\sigma_{r,\mathrm{out}}/\sqrt{5}$). In the presence of outliers, the median was a more reliable representative of the middle of the samples' distribution. In this case, the division factor $\sigma_{r,\mathrm{out}}/\sqrt{5}$ should be equal to the median of absolute deviations from the median \cite{Black1998,Rousseeuw1987}.
Accordingly, we derived $\sigma_{r,\mathrm{out}}/\sqrt{5}$ from the median of absolute deviations of the features differences from the median of the features differences. The features differences were calculated in the 26-connected neighborhoods of the samples. More specifically
\begin{subequations}
\label{eq:resOutRem}
\begin{equation}
\frac{\sigma_{r,\mathrm{out}}}{\sqrt{5}}=1.4826\times\median_{v_{r,j}\in\mathbb{D}_r}{\Big\|\mathbf{F}_{r,j}-\median_{v_{r,l}\in\mathbb{D}_r}\big(\mathbf{F}_{r,l}\big)\Big\|}_1
\end{equation}
\begin{equation}
\mathbf{F}_{r,j}={\big[\tilde{\mathbf{f}}_{r,j}-\tilde{\mathbf{f}}_{r,k}\big]}_{:,k}~~~~~~\forall v_{r,k}\in\mathbb{N}_{r,j}
\end{equation}
\begin{equation}
\mathrm{dim}(\mathbf{F}_{r,j})=\mathrm{dim}(\tilde{\mathbf{f}}_{r,j})\times 26
\end{equation}
\end{subequations}
with $1.4826$ being a normalization factor guaranteeing the equivalence of the standard deviation and the median of absolute deviations from the median. This way, each tuning parameter $\sigma_{r,\mathrm{out}}\in\mathbb{R}_{+}$ was specific to the samples $\mathbb{D}_r$ and the resolution layer $r\in\{0,\cdots,n_{\mathrm{lay}}\}$.

The proposed edge weights ${\{w_{r,jk}=w_{r,kj}\in\mathbb{R}_{\geq 0}\}}_{e_{r,jk}\in\mathbb{E}_{r,\mathrm{feats}}}$ given by \eqref{eq:weightTukeyFeats} underpinned an \textbf{implicit boundary detection capability} for the spatial feature-based subgraph $\mathcal{G}_{r,\mathrm{feats}}$. In addition, they enhanced the regularity or homogeneity of the subgraph $\mathcal{G}_{r,\mathrm{feats}}$ by preventing any weight from being vanished. This was due to the saturation of the Tukey's function at large features differences (gradients). These weights and the weights given by \eqref{eq:weightPriors2} for the aspatial prior-based subgraph $\mathcal{G}_{r,\mathrm{prior}}$ formed the combinatorial Dirichlet integral given by \eqref{eq:CombDirichInt2} for the graph $\mathcal{G}_r=\mathcal{G}_{r,\mathrm{feats}}\bigcup\mathcal{G}_{r,\mathrm{prior}}$. Then, the classification posterior $\hat{p}_{r,j,c}\in(0,1)$ estimated for each sample (vertex) $v_{r,j}\in\mathbb{V}_{r,\mathrm{feats}}$ with regard to each class $c\in\mathbb{L}$ should fulfill \eqref{eq:probEq2} on the graph $\mathcal{G}_r$. The optimum posteriors $\mathbf{P}_r={[\mathbf{p}_{r,c}]}_{:,c}={[\mathbf{p}_{r,j}]}_{j,:}={[p_{r,j,c}]}_{j,c}$ of the validation/test samples $\mathbb{D}_r=\mathbb{D}_{r,\mathrm{val}}\cup\mathbb{D}_{r,\mathrm{test}}$ of each resolution $r\in\{0,\cdots,n_{\mathrm{lay}}\}$ were the solutions of the system of sparse linear equations in \eqref{eq:linEqsUnlab2}.

\section{Constrained Feature- and Prior-based Graph}
\label{sec:CnstrFeatPriorGraph}
In a constrained feature- and prior-based graph, we further extended the spatial feature-based subgraph $\mathcal{G}_{r,\mathrm{feats}}$ of the graph $\mathcal{G}_r=\mathcal{G}_{r,\mathrm{feats}}\bigcup\mathcal{G}_{r,\mathrm{prior}}$ proposed in \autoref{sec:FeatPriorGraph} by
\begin{enumerate}[label={(\arabic*)},leftmargin=*]
\setlength\itemsep{0em}
\item explicitly detecting boundaries between different classes (e.g. objects) in spatial domain (e.g. image) by applying a 3D Sobel operator to each intensity channel of the samples (vertices);
\item categorizing the samples (vertices) according to their classification priors and the detected boundaries and reducing the number of unknown classifications by only focusing on challenging samples of low confident priors.
\end{enumerate}

For the subgraph $\mathcal{G}_{r,\mathrm{feats}}$, our proposed edge weights in \eqref{eq:weightTukeyFeats} implicitly detected boundaries between different classes in spatial domain. The explicit boundary detection proposed here could be considered as a supplement or as a replacement of that implicit boundary detection. In the latter case, the edge weights were given by \eqref{eq:weightFeats3}. Moreover, the categorization of the samples (vertices) produced some automatically labeled samples of assumed known posteriors. These prelabeled samples formed some constraints for the minimization of the combinatorial Dirichlet integral in \eqref{eq:CombDirichInt2}. They also reduced the computational complexity of the classifications by reducing the number of unknown posteriors and allowing to focus on challenging samples of low confident priors. This focus could also improve the optimization of the hyperparameter $\lambda_{r,\mathrm{prior}}$ introduced in \eqref{eq:weightPriors2} for the aspatial prior-based subgraph $\mathcal{G}_{r,\mathrm{prior}}$.
The low confident priors could stem from uncertainties in their estimation or errors in the manually labeled training samples used to train (optimize) the prior estimator.

The constrained graph $\mathcal{G}_r$ followed a one-vs-all classification approach. That is, for each important (foreground) class $c\in\mathbb{L}$ it categorized the samples (vertices) in $\mathbb{V}_{r,\mathrm{feats}}$ to:
\begin{equation}
\mathbb{V}_{r,\mathrm{fore},c}=\{v_{r,j}\in\mathbb{V}_{r,\mathrm{feats}}\big|a_{r,j,c}\geq 0.8\}
\label{eq:foreSet}
\end{equation}
\begin{equation}
\mathbb{V}_{r,\mathrm{back},c}=\{v_{r,j}\in\mathbb{V}_{r,\mathrm{feats}}\big|a_{r,j,c}\leq 0.2\}
\label{eq:backSet}
\end{equation}
\begin{equation}
\mathbb{V}_{r,\mathrm{hard},c}=~\text{Set~of~boundary~samples~detected~by~the~3D~Sobel~operator}
\label{eq:hardSet}
\end{equation}
\begin{equation}
\mathbb{V}_{r,\mathrm{soft},c}=\{v_{r,j}\in\mathbb{V}_{r,\mathrm{feats}}\big|a_{r,j,c}\in[0.4,0.6]\}
\label{eq:softSet}
\end{equation}
\begin{equation}
\mathbb{V}_{r,\mathrm{rest},c}=\mathbb{V}_{r,\mathrm{feats}}-(\mathbb{V}_{r,\mathrm{fore},c}\cup\mathbb{V}_{r,\mathrm{back},c}\cup\mathbb{V}_{r,\mathrm{hard},c})
\label{eq:restSet}
\end{equation}

\begin{figure}[t!]
\begin{center}
\includegraphics[width=0.9\textwidth]{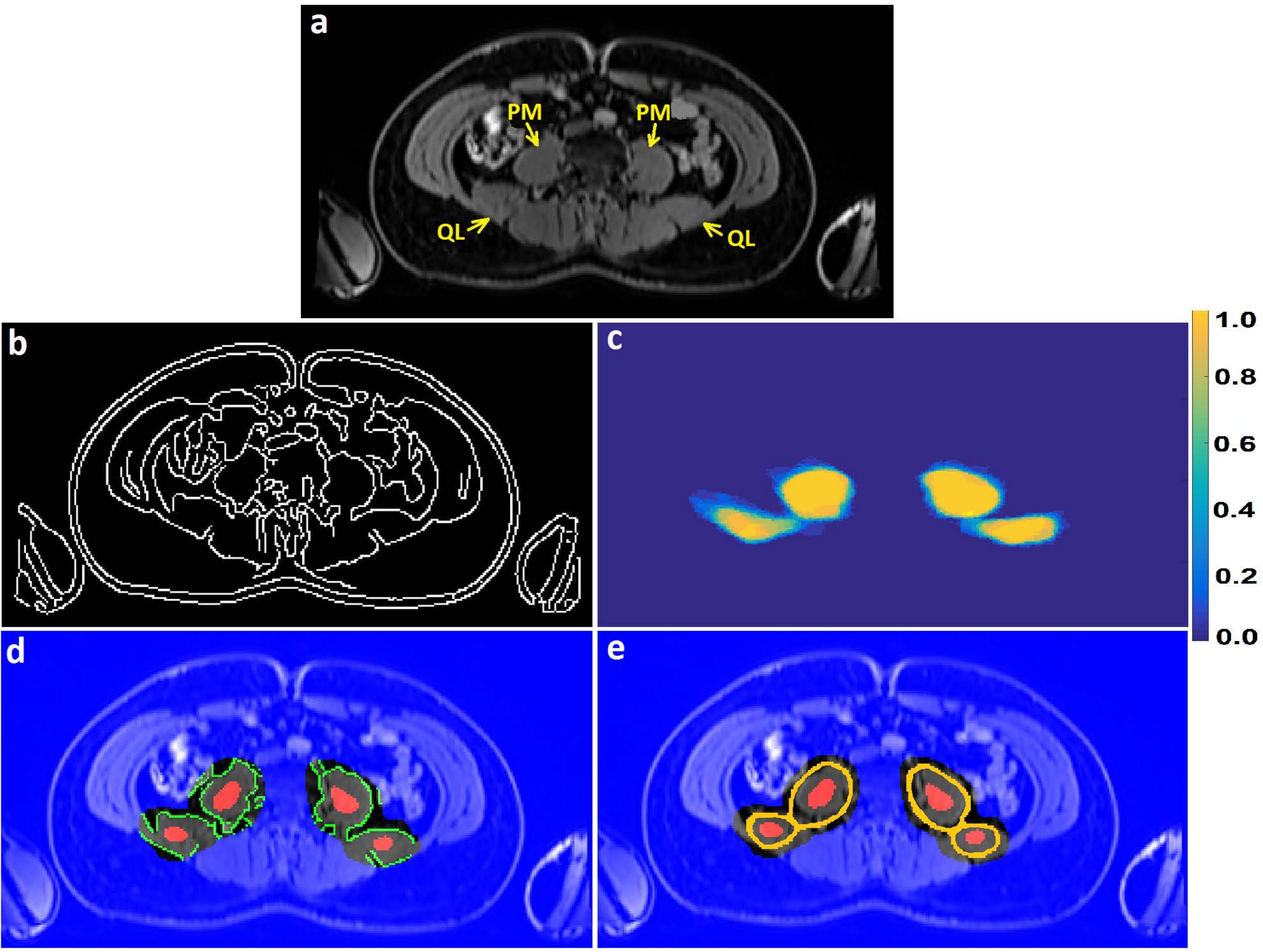}
\caption{a) An axial slice of a water image with the psoas major (PM) and quadratus lumborum (QL) muscles. b) Boundaries detected by the Sobel operator. c) Map of the priors estimated for the foreground (PM and QL muscles) class. d,e) Regions corresponding to $\mathbb{V}_{r,\mathrm{fore},c}$ (red), $\mathbb{V}_{r,\mathrm{back},c}$ (blue), $\mathbb{V}_{r,\mathrm{hard},c}$ (green), and $\mathbb{V}_{r,\mathrm{soft},c}$ (orange).}
\label{fig:VoxelSetsCnstrRW}
\end{center}
\end{figure}

\autoref{fig:VoxelSetsCnstrRW} shows these subsets on an axial slice of a water image with the important (foreground) class being the spatial regions of the psoas major and quadratus lumborum muscles.

For the aforementioned subsets, the condition in \eqref{eq:probEq2} boiled down to
\begin{equation}
\label{eq:probDirichCnstr}
\begin{split}
&\text{if}~v_{r,j}\in\mathbb{V}_{r,\mathrm{fore},c}:~~p_{r,j,c}=1~~~~~~~~\text{and}~~~~~~~~\text{if}~v_{r,j}\in\mathbb{V}_{r,\mathrm{back},c}:~~p_{r,j,c}=0\\
&\text{if}~v_{r,j}\in\mathbb{V}_{r,\mathrm{hard},c}:~~p_{r,j,c}=0.5\\
&\text{if}~v_{r,j}\in\mathbb{V}_{r,\mathrm{rest},c}:~~\hat{p}_{r,j,c}=\frac{\lambda_{r,\mathrm{prior}}\cdot a_{r,j,c}+\sum_{e_{r,jk}\in\mathbb{E}_{r,\mathrm{feats}}}w_{r,jk}\cdot\hat{p}_{r,k,c}}{d_{r,j}+\lambda_{r,\mathrm{prior}}}.
\end{split}
\end{equation}
That is, we assumed that the samples (vertices) in $\mathbb{V}_{r,\mathrm{fore},c}$, $\mathbb{V}_{r,\mathrm{back},c}$, and $\mathbb{V}_{r,\mathrm{hard},c}$ were prelabeled and had known classification posteriors. Accordingly, the combinatorial Dirichlet integral of the constrained graph was given by
\begin{subequations}
\label{eq:CombDirichCnstr}
\begin{align*}
\mathcal{L}_{\mathrm{Dirich}}(\hat{\mathbf{p}}_{r,c})&=\frac{1}{2}\Big[\sum_{\substack{e_{r,jk}\in\\\mathbb{E}_{r,\mathrm{feats}}}}w_{r,jk}\cdot(\hat{p}_{r,j,c}-\hat{p}_{r,k,c})^2+\sum_{\substack{v_{r,j}\in\\\mathbb{V}_{r,\mathrm{rest},c}}}\lambda_{r,\mathrm{prior}}\cdot(\hat{p}_{r,j,c}-a_{r,j,c})^2\Big]\\
&=\frac{1}{2}\underbrace{\begin{bmatrix}\hat{\mathbf{p}}_{r,c,\mathrm{lab}}^T&\hat{\mathbf{p}}_{r,c,\mathrm{unl}}^T\end{bmatrix}}_{\hat{\mathbf{p}}_{r,c}^T}\times\underbrace{\begin{bmatrix}\boldsymbol{\Gamma}_{r,\mathrm{lab}}&\boldsymbol{\Gamma}_{r,\mathrm{ula}}\\\boldsymbol{\Gamma}_{r,\mathrm{ula}}^T&\boldsymbol{\Gamma}_{r,\mathrm{unl}}\end{bmatrix}}_{\boldsymbol{\Gamma}_r}\times\underbrace{\begin{bmatrix}\hat{\mathbf{p}}_{r,c,\mathrm{lab}}\\\hat{\mathbf{p}}_{r,c,\mathrm{unl}}\end{bmatrix}}_{\hat{\mathbf{p}}_{r,c}}\tag{\ref{eq:CombDirichCnstr}}
\end{align*}
\end{subequations}
with the edge weights ${\{w_{r,jk}=w_{r,kj}\in\mathbb{R}_{\geq 0}\}}_{e_{r,jk}\in\mathbb{E}_{r,\mathrm{feats}}}$ being defined in \eqref{eq:weightTukeyFeats} (with implicit boundary detection) or \eqref{eq:weightFeats3} (without implicit boundary detecetion).

Then, the optimum classification posteriors $\mathbf{p}_{r,c}=\begin{bmatrix}\mathbf{p}_{r,c,\mathrm{lab}}\\\mathbf{p}_{r,c,\mathrm{unl}}\end{bmatrix}={[p_{r,j,c}]}_j$ were given by
\begin{equation}
\label{eq:DirichCnstr}
\mathbf{p}_{r,c}=\argmin_{\hat{\mathbf{p}}_{r,c}}~{\mathcal{L}_{\mathrm{Dirich}}(\hat{\mathbf{p}}_{r,c})}~~~\text{subject~to}:\begin{cases}
\forall v_{r,j}\in\mathbb{V}_{r,\mathrm{fore},c}:~&p_{r,j,c}=1\\
\forall v_{r,j}\in\mathbb{V}_{r,\mathrm{back},c}:~&p_{r,j,c}=0\\
\forall v_{r,j}\in\mathbb{V}_{r,\mathrm{hard},c}:~&p_{r,j,c}=0.5
\end{cases}.
\end{equation}

These posteriors were the solutions of a system of sparse linear equations given by \eqref{eq:linEqsRandWalk2} as
\begin{equation}
\label{eq:linEqsCnstr}
\boldsymbol{\Gamma}_{r,\mathrm{unl}}\times\mathbf{p}_{r,c,\mathrm{unl}}=-\boldsymbol{\Gamma}_{r,\mathrm{ula}}^T\times\mathbf{p}_{r,c,\mathrm{lab}}+\lambda_{r,\mathrm{prior}}\cdot\mathbf{a}_{r,c,\mathrm{unl}}
\end{equation}
with $\mathbb{V}_{r,\mathrm{lab}}=\mathbb{V}_{r,\mathrm{fore},c}\cup\mathbb{V}_{r,\mathrm{back},c}\cup\mathbb{V}_{r,\mathrm{hard},c}$ and $\mathbb{V}_{r,\mathrm{unl}}=\mathbb{V}_{r,\mathrm{rest},c}$.

\section{Guided Feature- and Prior-based Graph}
\label{sec:GuidedFeatPriorGraph}
\begin{figure}[t!]
\begin{center}
\includegraphics[width=1.0\textwidth]{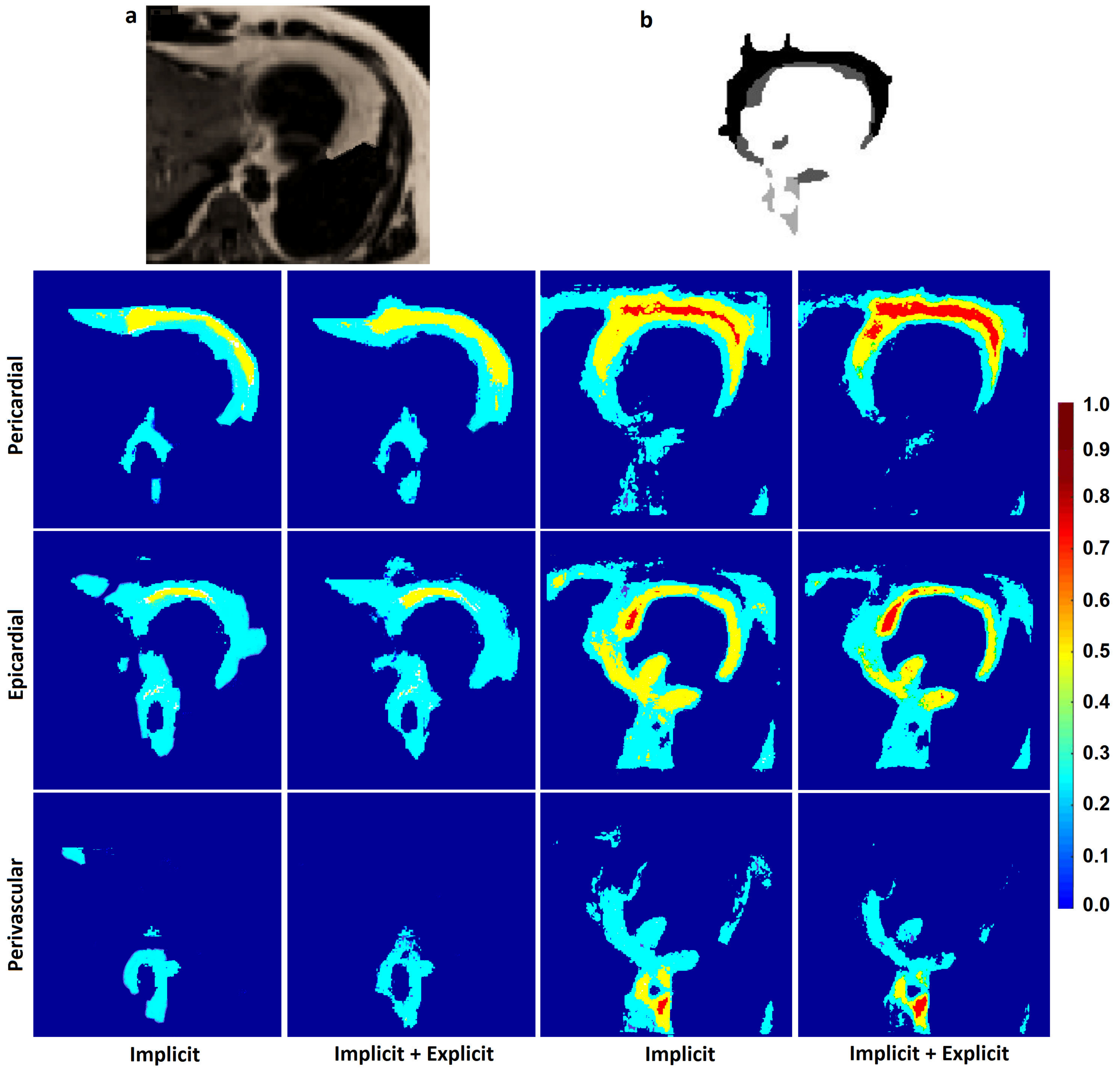}
\caption{An axial slice of a fat image (a), an axial slice of manually annotated cardiac adipose tissues (b), and the voxelwise classification posteriors estimated for these tissues by the implicit and/or explicit boundary detection in \autoref{sec:FeatPriorGraph} and \autoref{sec:CnstrFeatPriorGraph}.}
\label{fig:PostCardFatImpExp}
\end{center}
\end{figure}

The graphs proposed in \autoref{sec:FeatPriorGraph} and \autoref{sec:CnstrFeatPriorGraph} tried to implicitly or explicitly detect boundaries between different classes (e.g. objects) in spatial domain (e.g. image). The implicit boundary detection used the Tukey's function of the features differences of the connected samples and the explicit boundary detection applied a 3D Sobel operator to each intensity channel of the samples. Both of these techniques could enhance the accuracies of a classification in spatial domain. However, their performance was limited when the features of different classes were similar or the boundaries were too weak to be detected. An example of these cases was segmentation of cardiac adipose tissues on fat-water MR images. As shown in \autoref{fig:PriorProbCAT} and \autoref{fig:AxialSegCAT} the random forest classifier proposed in \cite{Fallah2018a,Fallah2018p,Fallah2019a,FallahJ2019} could only detect the overall adipose tissues but could not differentiate different kinds of it. The neighborhood graph $\mathcal{G}_r=\mathcal{G}_{r,\mathrm{feats}}\bigcup\mathcal{G}_{r,\mathrm{prior}}$ proposed in \autoref{sec:FeatPriorGraph} or \autoref{sec:CnstrFeatPriorGraph} took the priors, the features, and the classification reliabilities of the samples and applied an implicit or explicit boundary detection. This could enhance the accuracy of the segmentation of the cardiac adipose tissues on fat-water MR images but the performance was still unsatisfactory. \autoref{fig:PostCardFatImpExp} shows the voxelwise classification posteriors estimated for these adipose tissues by the implicit and/or explicit boundary detection on two axial slices of a fat image.
To tackle features similarities and undetectable boundaries between different classes in spatial domain, we further extended the spatial feature-based subgraph $\mathcal{G}_{r,\mathrm{feats}}$ by guiding the classification through additional information. These information got encoded into a diffusion-based susceptible-infected-recovered (SIR) model proposed in \cite{Bampis2017}.

\subsection{Susceptible-Infected-Recovered (SIR) Model}
\label{ssec:SirModel}
The SIR model was a temporal model of infections' diffusions in a population whose inter-personal contacts were represented by a neighborhood graph. Thus, the neighborhood graph $\mathcal{G}_r=\mathcal{G}_{r,\mathrm{feats}}\bigcup\mathcal{G}_{r,\mathrm{prior}}$ proposed in \autoref{sec:FeatPriorGraph} and \autoref{sec:CnstrFeatPriorGraph} could accommodate this model in spatial domain. For sake of clarity, we initially disconnected the aspatial prior-based subgraph $\mathcal{G}_{r,\mathrm{prior}}$ from the spatial feature-based subgraph $\mathcal{G}_{r,\mathrm{feats}}$ by setting $\lambda_{r,\mathrm{prior}}=0$. This converted the condition in \eqref{eq:probEq2} for each classification posterior $\hat{p}_{r,j,c}\in(0,1)$ to
\begin{equation}
\hat{p}_{r,j,c}=\frac{1}{d_{r,j}}\sum_{e_{r,jk}\in\mathbb{E}_{r,\mathrm{feats}}}w_{r,jk}\cdot\hat{p}_{r,k,c}.
\label{eq:probEq3}
\end{equation}

In the SIR model, every sample (vertex) $v_{r,j}\in\mathbb{V}_{r,\mathrm{feats}}$ got handled like a person whose contacts were limited to its direct spatial neighbors on the graph $\mathcal{G}_r$. Also, each infection was represented by a class $c\in\mathbb{L}$. Then, the probability that a person (sample) $v_{r,j}\in\mathbb{V}_{r,\mathrm{feats}}$ got infected by an infection (class) $c\in\mathbb{L}$ at a time $t\in[0,\infty)$ was $\hat{p}_{r,j,c}^{(t)}\in(0,1)$. Accordingly, the temporal evolution of each infection $c\in\mathbb{L}$ for each person (sample) $v_{r,j}\in\mathbb{V}_{r,\mathrm{feats}}$ was
\begin{equation}
\Delta\hat{p}_{r,j,c}^{(t)}=\hat{p}_{r,j,c}^{(t+\Delta t)}-\hat{p}_{r,j,c}^{(t)}~~\text{with}~~\lim_{t\to+\infty}\hat{p}_{r,j,c}^{(t)}=\hat{p}_{r,j,c}^{(t_{\infty})}~~\text{and}~~\Delta\hat{p}_{r,j,c}^{(t_{\infty})}=0.
\label{eq:tempEvolInf}
\end{equation}

The SIR model assumed that every person (sample) had no recovery state but a constant susceptibility to every infection (class). It also assumed that all the persons (samples) had the same susceptibility to all the infections (classes) \cite{Bampis2017}. In contrast, we assumed that each person (sample) $v_{r,j}\in\mathbb{V}_{r,\mathrm{feats}}$ had a specific susceptibility $s_{r,j,c}\in(0,1)$ to each infection (class) $c\in\mathbb{L}$. That is, each susceptibility $s_{r,j,c}\in(0,1)$ was \textbf{sample- and class-specific}. This allowed us to encode additional information based on the characteristics of the sample $v_{r,j}\in\mathbb{V}_{r,\mathrm{feats}}$ and the class $c\in\mathbb{L}$ into each susceptibility $s_{r,j,c}\in(0,1)$.
According to a locally averaged SIR model, the temporal derivative of the evolution of each infection (class) $c\in\mathbb{L}$ for each person (sample) $v_{r,j}\in\mathbb{V}_{r,\mathrm{feats}}$ was a \textbf{weighted average} of the relative infection probabilities of the sample and its direct spatial neighbors on the subgraph $\mathcal{G}_{r,\mathrm{feats}}=(\mathbb{V}_{r,\mathrm{feats}},\mathbb{E}_{r,\mathrm{feats}})$. The weights of the averaging were formed from the susceptibilities of the sample and its neighbors as well as the original edge weights of the subgraph $\mathcal{G}_{r,\mathrm{feats}}$ connecting the sample to its neighbors \cite{Postnikov2007}. More specifically,
\begin{equation}
\frac{\Delta \hat{p}_{r,j,c}^{(t)}}{\Delta t}=\sum_{e_{r,jk}\in\mathbb{E}_{r,\mathrm{feats}}}\frac{w_{r,jk}}{\sqrt{d_{r,j}}}\cdot(\frac{s_{r,k,c}}{\sqrt{d_{r,k}}}\hat{p}_{r,k,c}^{(t)}-\frac{s_{r,j,c}}{\sqrt{d_{r,j}}}\hat{p}_{r,j,c}^{(t)})
\label{eq:infinEvolInf}
\end{equation}
with $w_{r,jk}\in\mathbb{R}_{\geq 0}$ being the weight given by \eqref{eq:weightFeats3} or \eqref{eq:weightTukeyFeats} for the edge $e_{r,jk}\in\mathbb{E}_{r,\mathrm{feats}}$ and $d_{r,j}=\sum_{e_{r,jk}\in\mathbb{E}_{r,\mathrm{feats}}}w_{r,jk}$. Then, by assuming $\Delta t=1$ and $\Delta\hat{p}_{r,j,c}^{(t_{\infty})}=0$, one obtained
\begin{equation}
\label{eq:steadyInfection}
\hat{p}_{r,j,c}^{(t_{\infty})}=\frac{1}{\sqrt{d_{r,j}}}\sum_{\substack{e_{r,jk}\in\\\mathbb{E}_{r,\mathrm{feats}}}}~\underbrace{\frac{w_{r,jk}}{\sqrt{d_{r,k}}}\cdot\frac{s_{r,k,c}}{s_{r,j,c}}}_{\substack{\text{modified}\\\text{weight}}~=~w'_{r,jk}}\cdot~\hat{p}_{r,k,c}^{(t_{\infty})}=\frac{1}{\sqrt{d_{r,j}}}\sum_{\substack{e_{r,jk}\in\\\mathbb{E}_{r,\mathrm{feats}}}}w'_{r,jk}\cdot\hat{p}_{r,k,c}^{(t_{\infty})}.
\end{equation}

That is, in the steady state $t_{\infty}$, the infection probability $\hat{p}_{r,j,c}^{(t_{\infty})}\in(0,1)$ of each person (sample) $v_{r,j}\in\mathbb{V}_{r,\mathrm{feats}}$ was a weighted average of the infection probabilities of its direct spatial neighbors. This condition and the comparison of \eqref{eq:steadyInfection} and \eqref{eq:probEq3} revealed that $\hat{p}_{r,j,c}^{(t_{\infty})}\in(0,1)$ was equivalent to the classification posterior $\hat{p}_{r,j,c}\in(0,1)$ of $v_{r,j}\in\mathbb{V}_{r,\mathrm{feats}}$ if
\begin{equation}
\label{eq:steadyInfectCond}
\sqrt{d_{r,j}}=d'_{r,j}=\sum_{e_{r,jk}\in\mathbb{E}_{r,\mathrm{feats}}}w'_{r,jk}=\sum_{e_{r,jk}\in\mathbb{E}_{r,\mathrm{feats}}}\frac{w_{r,jk}}{\sqrt{d_{r,k}}}\cdot\frac{s_{r,k,c}}{s_{r,j,c}}.
\end{equation}

The only difference was that $\hat{p}_{r,j,c}^{(t_{\infty})}\in(0,1)$ involved ${\{w'_{r,jk}\in\mathbb{R}_{\geq 0}\}}_{e_{r,jk}\in\mathbb{E}_{r,\mathrm{feats}}}$ and $\hat{p}_{r,j,c}\in(0,1)$ involved ${\{w_{r,jk}\in\mathbb{R}_{\geq 0}\}}_{e_{r,jk}\in\mathbb{E}_{r,\mathrm{feats}}}$. Each $w_{r,jk}\in\mathbb{R}_{\geq 0}$ encoded features differences of the connected samples $v_{r,j}\in\mathbb{V}_{r,\mathrm{feats}}$ and $v_{r,k}\in\mathbb{V}_{r,\mathrm{feats}}$. However, each $w'_{r,jk}=(w_{r,jk}/\sqrt{d_{r,k}})\cdot(s_{r,k,c}/s_{r,j,c})$ encoded the aforementioned features differences in $w_{r,jk}\in\mathbb{R}_{\geq 0}$, the overall features differences of $v_{r,k}\in\mathbb{V}_{r,\mathrm{feats}}$ and its direct spatial neighbors in $\sqrt{d_{r,k}}\in\mathbb{R}_{+}$, and some other (\textbf{application-specific}) information in $s_{r,j,c}\in(0,1)$ and $s_{r,k,c}\in(0,1)$.

As described in \autoref{sec:OutlineGraphs}, each sample (vertex) had 26 direct spatial neighbors in a 26-connected neighborhood which was a second order neighborhood in 3D. In this regard, the subgraph $\mathcal{G}_{r,\mathrm{feats}}$ formed from the edge weights ${\{w_{r,jk}\}}_{e_{r,jk}\in\mathbb{E}_{r,\mathrm{feats}}}$ was a graph of \textbf{second order} neighborhoods. However, by including $\sqrt{d_{r,k}}\in\mathbb{R}_{+}$ into each modified edge weight $w'_{r,jk}\in\mathbb{R}_{\geq 0}$, the subgraph $\mathcal{G}_{r,\mathrm{feats}}$ formed from ${\{w'_{r,jk}\}}_{e_{r,jk}\in\mathbb{E}_{r,\mathrm{feats}}}$ become a graph of \textbf{infinite order} neighborhoods. This was because, as given by \eqref{eq:steadyInfectCond}, each $d_{r,j}\in\mathbb{R}_{+}$ (and similarly each $d_{r,k}\in\mathbb{R}_{+}$) was a consolidation of features differences of second order neighborhoods (through $w_{r,jk}\in\mathbb{R}_{\geq 0}$) as well as features differences of second order neighborhoods of second order neighborhoods and so on (through $\sqrt{d_{r,k}}\in\mathbb{R}_{+}$). The equivalence of $\hat{p}_{r,j,c}^{(t_{\infty})}\in(0,1)$ and $\hat{p}_{r,j,c}\in(0,1)$ allowed to denote $\hat{p}'_{r,j,c}=\hat{p}_{r,j,c}^{(t_{\infty})}\in(0,1)$ and also reconnect the aspatial prior-based subgraph $\mathcal{G}_{r,\mathrm{prior}}$ to the spatial feature-based subgraph $\mathcal{G}_{r,\mathrm{feats}}$ by setting $\lambda_{r,\mathrm{prior}}\neq 0$. This reconnection resulted in
\begin{equation}
\hat{p}'_{r,j,c}=\frac{\lambda_{r,\mathrm{prior}}\cdot a_{r,j,c}+\sum_{e_{r,jk}\in\mathbb{E}_{r,\mathrm{feats}}}w'_{r,jk}\cdot\hat{p}'_{r,k,c}}{d'_{r,j}+\lambda_{r,\mathrm{prior}}}
\label{eq:probEqFeatPrior4}
\end{equation}
to be fulfilled by each infection probability $\hat{p}'_{r,j,c}\in(0,1)$ on the graph $\mathcal{G}_r=\mathcal{G}_{r,\mathrm{feats}}\bigcup\mathcal{G}_{r,\mathrm{prior}}$.

The condition \eqref{eq:probEqFeatPrior4} was equivalent to the condition \eqref{eq:probEq2} for the classification posterior $\hat{p}_{r,j,c}\in(0,1)$ and thus resulted in the following Dirichlet integral over the graph $\mathcal{G}_r$:
\begin{subequations}
\label{eq:CombDirichIntSIR}
\begin{align*}
\mathcal{L}_{\mathrm{Dirich}}(\hat{\mathbf{P}}'_r)&=\frac{1}{2}{\hat{\mathbf{P}}}^{\prime^T}_r\times\boldsymbol{\Gamma}'_r\times\hat{\mathbf{P}}'_r\tag{\ref{eq:CombDirichIntSIR}}\\
&=\frac{1}{2}\sum_{c=1}^{n_{\mathrm{clas}}-1}\Bigg[\sum_{\substack{e_{r,jk}\in\\\mathbb{E}_{r,\mathrm{feats}}}}w'_{r,jk}\cdot(\hat{p}'_{r,j,c}-\hat{p}'_{r,k,c})^2+\sum_{\substack{v_{r,j}\in\\\mathbb{V}_{r,\mathrm{feats}}}}\lambda_{r,\mathrm{prior}}\cdot(\hat{p}'_{r,j,c}-a_{r,j,c})^2\Bigg].
\end{align*}
\end{subequations}
Here, it was assumed that all the samples were unlabeled, i.e. $\mathbb{V}_{r,\mathrm{unl}}=\mathbb{V}_{r,\mathrm{feats}}$ and $\mathbb{D}_{r,\mathrm{unl}}=\mathbb{D}_r$; $\hat{\mathbf{P}}'_r={[\hat{\mathbf{p}}'_{r,c}]}_{:,c}={[\hat{\mathbf{p}}'_{r,j}]}_{j,:}={[\hat{p}'_{r,j,c}]}_{j,c}$ was the $|\mathbb{D}_r|\times n_{\mathrm{clas}}$ matrix of the estimated classification posteriors (infection probabilities) and $\boldsymbol{\Gamma}'_r$ was the combinatorial Laplace-Beltrami matrix of the guided graph $\mathcal{G}_r=\mathcal{G}_{r,\mathrm{feats}}\bigcup\mathcal{G}_{r,\mathrm{prior}}$ given by
\begin{equation}
\label{eq:combLapMtrx3}
\gamma'_{r,j,k}=\begin{cases}
d'_{r,j}+\lambda_{r,\mathrm{prior}}&\text{if}~v_{r,j}=v_{r,k}\\
-w'_{r,jk}&\text{if}~e_{r,jk}\in\mathbb{E}_{r,\mathrm{feats}}\\
0&\text{otherwise}
\end{cases}~~~~~\boldsymbol{\Gamma}'_r={[\gamma'_{r,j,k}]}_{j,k}.
\end{equation}
Then, the optimum classification posteriors $\mathbf{P}'_r={[\mathbf{p}'_{r,c}]}_{:,c}={[\mathbf{p}'_{r,j}]}_{j,:}={[p'_{r,j,c}]}_{j,c}$ were
\begin{equation}
\label{eq:DirichMin3}
\mathbf{P}'_r=\argmin_{\hat{\mathbf{P}}'_r}~\mathcal{L}_{\mathrm{Dirich}}(\hat{\mathbf{P}}'_r).
\end{equation}
These posteriors were the solutions of \eqref{eq:linEqsUnlab2} with $\boldsymbol{\Gamma}_r$ being replaced with $\boldsymbol{\Gamma}'_r$.

The SIR model could also be encoded into the constrained graph proposed in \autoref{sec:CnstrFeatPriorGraph}. This resulted in the combinatorial Dirichlet integral given by \eqref{eq:CombDirichCnstr}, the optimum classification posteriors $\mathbf{p}'_{r,c}={[p'_{r,j,c}]}_j$ given by \eqref{eq:DirichCnstr}, and the system of equations given by \eqref{eq:linEqsCnstr} with $\boldsymbol{\Gamma}_r$ and $\mathbf{p}_{r,c}={[p_{r,j,c}]}_j$ being replaced with $\boldsymbol{\Gamma}'_r$ and $\mathbf{p}'_{r,c}={[p'_{r,j,c}]}_j$ respectively.

\begin{figure}[t!]
\begin{center}
\includegraphics[width=1.0\textwidth]{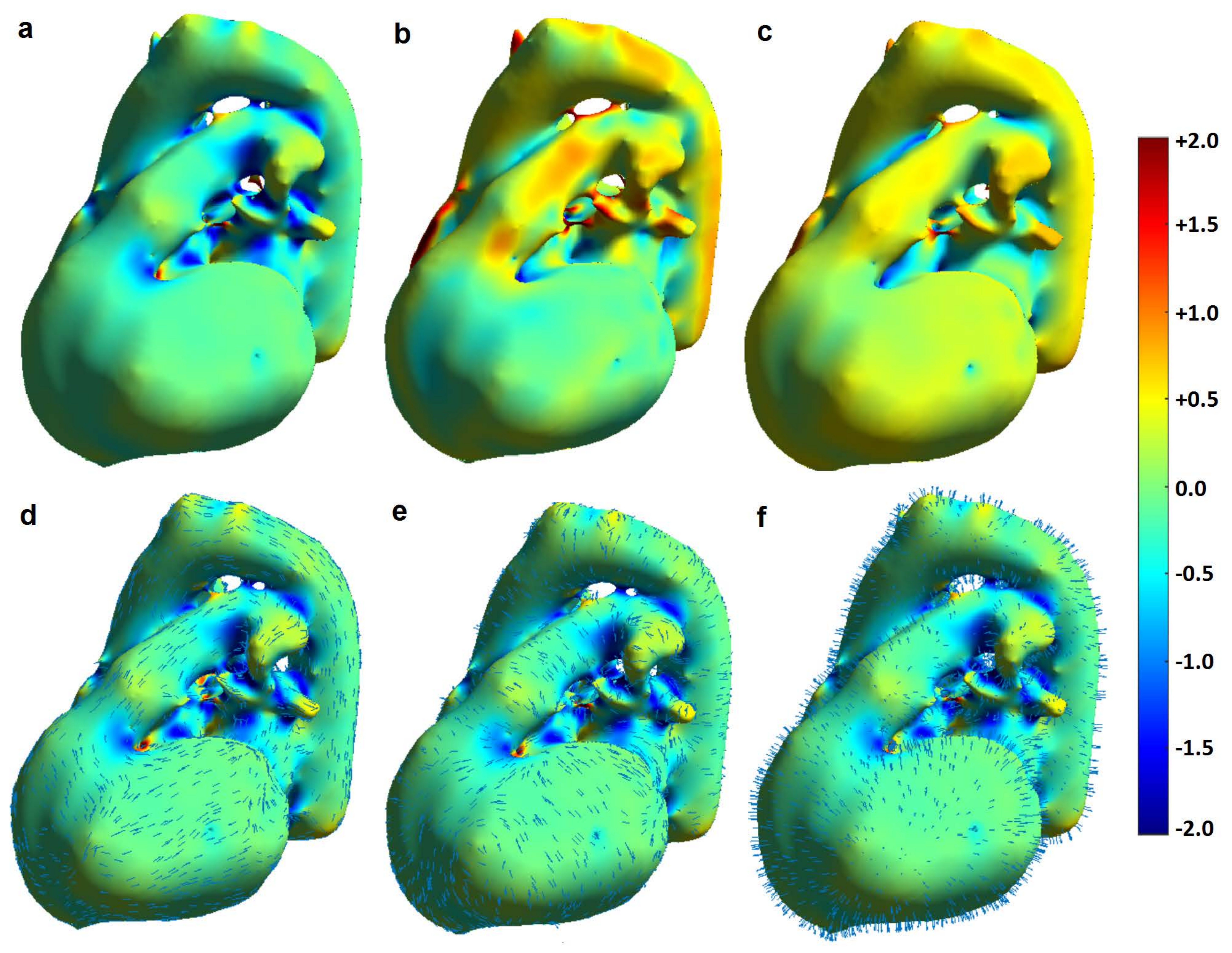}
\caption{The color-coded Gaussian (a), maximum (b), and mean (c) curvatures with the directions of minimum (d) and maximum (e) curvatures and normals (f) of the surface of a cardiac structure segmented over a volumetric water image. The color-coding was clipped to [-2,+2] for a better visualization.}
\label{fig:CurvsCardStruct}
\end{center}
\end{figure}

\subsection{Derivation of Susceptibilities}
\label{ssec:SampClasSuscpt}
As mentioned earlier, each susceptibility $s_{r,j,c}\in(0,1)$ was sample-, class-, and application-specific. In the following, we derived these susceptibilities for \textbf{segmenting cardiac adipose tissues on a fat-water MR image} by using information from the fat image and the surface of the cardiac structure segmented over the corresponding water image. The cardiac structure was composed of myocardium and cardiac vessels. These lean tissues were mainly composed of water and thus appeared bright on the water images. The cardiac adipose tissues included epicardial (EpAT), pericardial (PeAT), and cardiac perivascular adipose tissue (PvAT). That is, $\mathbb{L}=\{\mathrm{EpAT},\mathrm{PeAT},\mathrm{PvAT},\mathrm{Background}\}=\{1,2,3,4\}$ and $n_{\mathrm{clas}}=4$. If a surface was smooth and $\mathbf{e}_{\mathrm{uvp}}\in\mathbb{R}^2$ was a unit vector in the tangent space\footnote{The tangent space of a $d$-dimensional Riemannian manifold at a point on the manifold was a $d$-dimensional vector space containing all the vectors tangent to the manifold at the point.} of the surface at a point, then normal curvature $\kappa_{\mathrm{n}}$ of the surface at the point and along the vector $\mathbf{e}_{\mathrm{uvp}}$ was inverse of the radius of the circle that best fitted the surface at the point and along $\mathbf{e}_{\mathrm{uvp}}$. With $\underline{\mathbf{C}}_{\mathrm{scp}}$ being a symmetric tensor expressing all the surface curvatures with regard to the local isometry of the surface to a (flat) Euclidean space at the point, $\kappa_{\mathrm{n}}$ was given by
\begin{equation}
\label{eq:normalCurvSmooth}
\kappa_{\mathrm{n}}=\mathbf{e}_{\mathrm{uvp}}^T\times\underline{\mathbf{C}}_{\mathrm{scp}}\cdot\mathbf{e}_{\mathrm{uvp}}.
\end{equation}

The orthonormal coordinate system of each tangent space could be rotated in such a way that it led to a diagonalized curvature tensor. In addition, if the surface was parametrized with 2 parameters\footnote{That is, the surface was a two-dimensional Riemannian manifold.}, then every tangent space on it become a tangent plane and the curvature tensor at each point become a curvature matrix \cite{Rusinkiewicz2004}. In these cases, we got
\begin{equation}
\label{eq:diagNormalCurvSmooth}
\kappa_{\mathrm{n}}=\mathbf{e}_{\mathrm{uvp}}^{'T}\cdot\begin{pmatrix}\kappa_{\mathrm{min}}&0\\0&\kappa_{\mathrm{max}}\end{pmatrix}\cdot\mathbf{e}'_{\mathrm{uvp}}=\begin{pmatrix}e'_1&e'_2\end{pmatrix}\cdot\begin{pmatrix}\kappa_{\mathrm{min}}&0\\0&\kappa_{\mathrm{max}}\end{pmatrix}\cdot\begin{pmatrix}e'_1\\e'_2\end{pmatrix}=\kappa_{\mathrm{min}}\cdot e_1^{'2}+\kappa_{\mathrm{max}}\cdot e_2^{'2}
\end{equation}
where $\kappa_{\mathrm{min}}$ and $\kappa_{\mathrm{max}}$ were the minimum and maximum curvatures of the surface, respectively, and $\mathbf{e}'_{\mathrm{uvp}}$ was another unit vector in the same tangent plane. This implied that the curvature matrix had eigenvalues of $\kappa_{\mathrm{min}}$, $\kappa_{\mathrm{max}}$, and $0$ with the respective eigenvectors of $\boldsymbol{\kappa}_{\mathrm{min}}$, $\boldsymbol{\kappa}_{\mathrm{max}}$, and $\boldsymbol{\kappa}_{\mathrm{norm}}$. Here, $\boldsymbol{\kappa}_{\mathrm{min}}$ and $\boldsymbol{\kappa}_{\mathrm{max}}$ denoted the directions of the minimum and maximum curvatures, respectively, and $\boldsymbol{\kappa}_{\mathrm{norm}}$ was the surface normal at the point. From these, the Gaussian, $\kappa_{\mathrm{gauss}}=\kappa_{\mathrm{min}}\cdot\kappa_{\mathrm{max}}$, and the mean, $\kappa_{\mathrm{mean}}=(\kappa_{\mathrm{min}}+\kappa_{\mathrm{max}})/2$, curvatures could be obtained. A point on the surface was planar if $\kappa_{\mathrm{gauss}}=\kappa_{\mathrm{mean}}=0$, elliptic if $\kappa_{\mathrm{gauss}}>0$, hyperbolic if $\kappa_{\mathrm{gauss}}<0$, and parabolic if $\kappa_{\mathrm{gauss}}=0~\text{and}~\kappa_{\mathrm{mean}}\neq 0$ \cite{Gatzke2006,Rusinkiewicz2004}.

\autoref{fig:CurvsCardStruct} shows color-coded curvatures and directions of normals and minimum and maximum curvatures of the surface of a cardiac structure segmented over a volumetric water image. Among the introduced curvatures, the maximum curvature had the maximum variance and could thereby capture most of the shape variations over a surface.

\begin{table}[t!]
\begin{center}
\caption{Number of voxels per patch at different resolutions of the pyramid with $n_{\mathrm{lay}}=5$ layers.}
\vspace{1mm}
\label{table:numberOfVoxelPerPatch}
\resizebox{\textwidth}{!}{%
\begin{tabular}{|c|c|c|c|c|c|c|}
\hline
\textbf{Resolution Layer (r)}&5&4&3&2&1&0\\\hline
\#\textbf{Voxels per Patch}&$48^3=110592$&$24^3=13824$&$12^3=1728$&$6^3=216$&$3^3=27$&$1$\\\hline
\end{tabular}}
\end{center}
\end{table}

The multiresolution fat-water patches of the multiresolution validation/test samples ${\{\mathbb{D}_r\}}_{r=0}^{n_{\mathrm{lay}}}$ got extracted from the addressed fat-water image according to the pyramid described in \cite{Fallah2018a,Fallah2018p,Fallah2019a,FallahJ2019}. In this pyramid, each patch at the resolution $r\in\{1,\cdots,n_{\mathrm{lay}}\}$ had $\big(3\times2^{r-1}\big)^3$ voxels and each patch at the resolution $r=0$ had one voxel. \autoref{table:numberOfVoxelPerPatch} shows the number of voxels per patch at different resolutions of the pyramid.

During our investigation of the features which could support a classification of the multiresolution fat-water patches, we found that modes of the histogram of oriented gradients (HOGs) of the fat channels of the samples of each resolution in the region of the cardiac adipose tissues had specific alignments with the maximum curvatures of the nearby surface of the cardiac structure. In particular,
\begin{itemize}[leftmargin=*]
\item modes of the HOGs of the fat channels of the samples of epicardial adipose tissue (EpAT) were aligned with the directions of the maximum curvatures of the myocardium;
\item modes of the HOGs of the fat channels of the samples of perivascular adipose tissue (PvAT) were aligned with the directions of the maximum curvatures of the cardiac vessels;
\item modes of the HOGs of the fat channels of the samples of pericardial adipose tissue (PeAT) were diverse without any specific alignment.
\end{itemize}

\begin{figure}[t!]
\begin{center}
\includegraphics[width=1.0\textwidth,height=18.5cm]{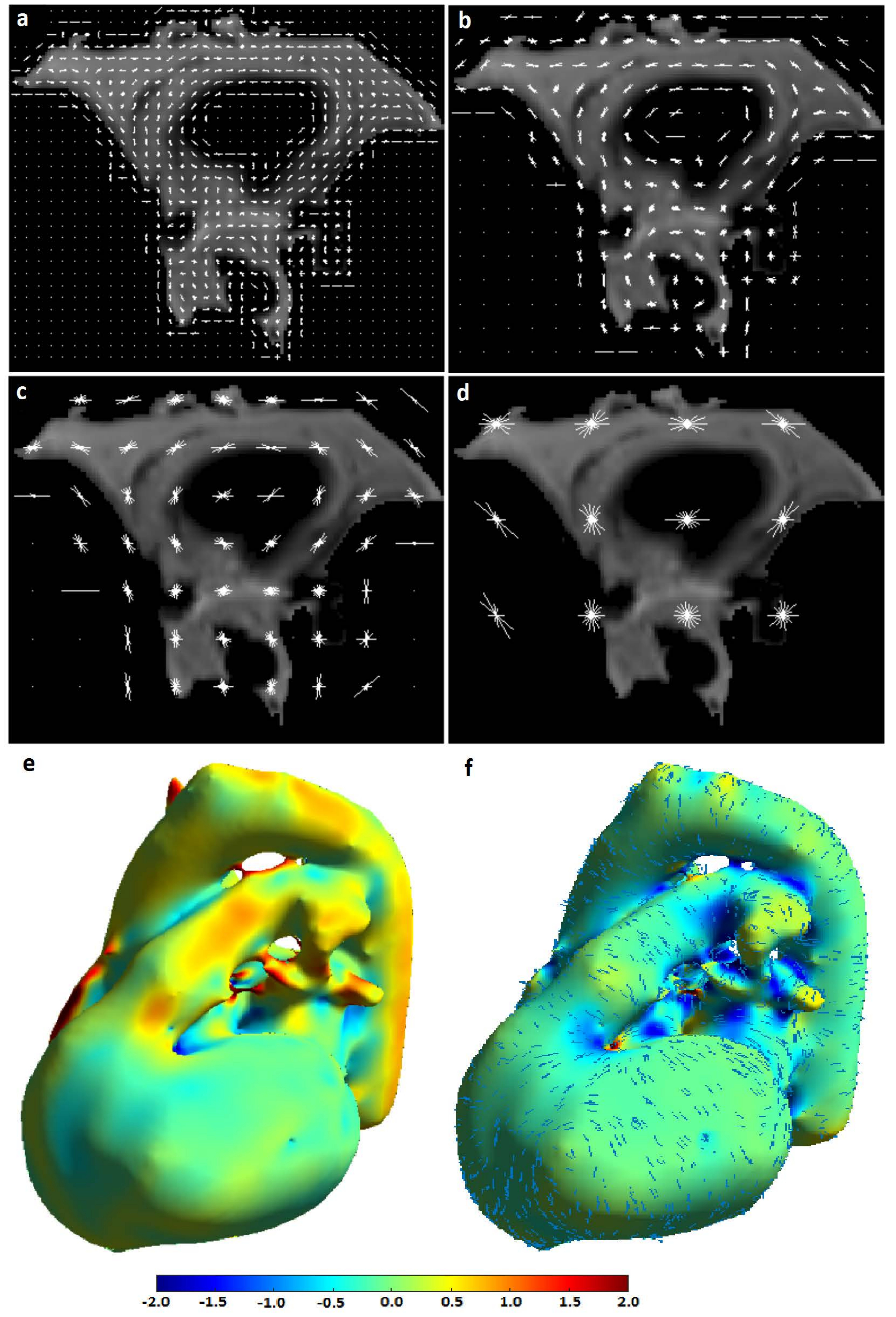}
\caption{HOGs of the fat channels of the samples at the second (a), third (b), fourth (c), and fifth (d) resolution layers of the pyramid shown on an axial slice of a fat image. The color-coded maximum (e) and Gaussian (f) curvatures with the directions of the maximum curvatures (f) of the cardiac structure segmented on the corresponding water image.}
\label{fig:ModesOfHogsCATs}
\end{center}
\end{figure}

These alignments motivated us to encode the \textbf{maximum curvatures of the surface of the cardiac structure} into the susceptibilities of the samples (vertices) over the graph $\mathcal{G}_r$. These susceptibilities provided additional information and formed the SIR model, described in \autoref{ssec:SirModel}, for segmenting cardiac adipose tissues on a fat-water MR image.

The aforementioned alignments could be attributed to the synchronous movement of the EpAT with the myocardium during the cardiac motion and the gradual development of the PvAT around the cardiac vessels. \autoref{fig:ModesOfHogsCATs} shows that these alignments were more or less observed in samples (image patches) of every resolution of the pyramid.
Use of the surface curvatures implied to segment the cardiac structure on the water channel of the addressed volumetric fat-water image and then compute its surface. This segmentation and surface computation was described in \autoref{ssec:SegCardiacStruct}.

\begin{figure}[t!]
\begin{center}
\includegraphics[width=1.0\textwidth]{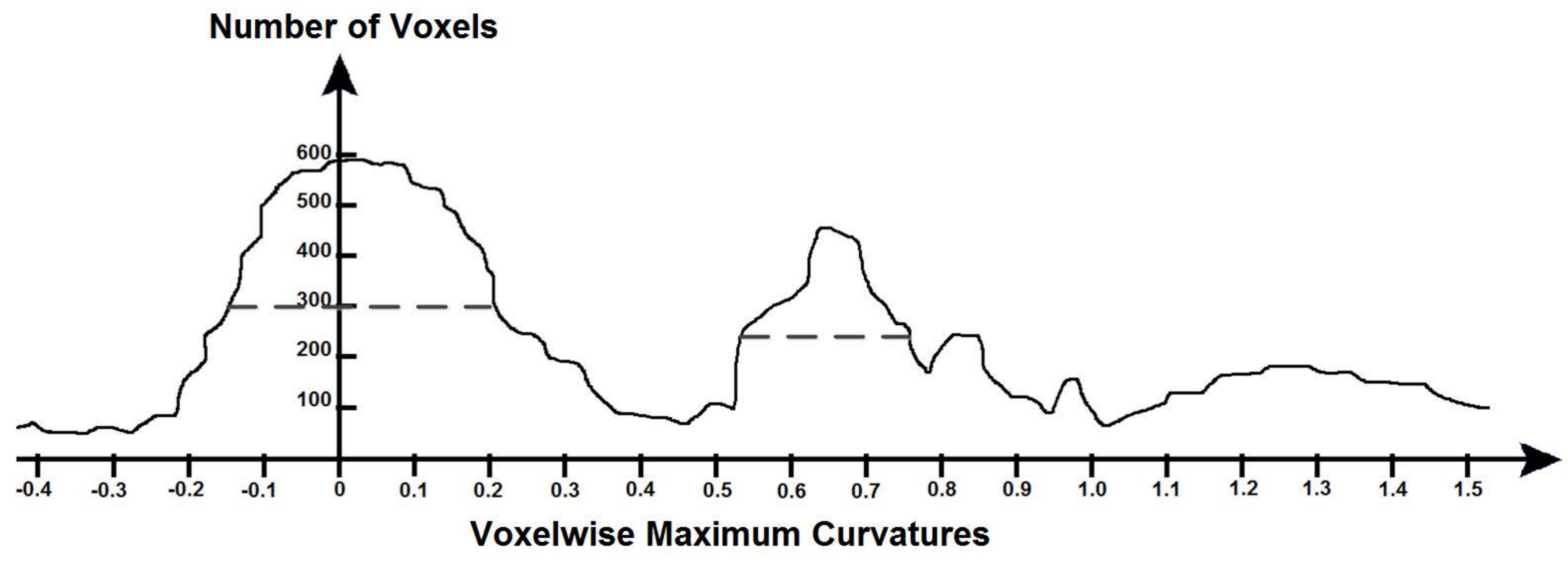}
\caption{Histogram of the magnitudes of the maximum curvatures of the surface voxels of a cardiac structure segmented over a water image and the FWHMs of its main peaks.}
\label{fig:HistCurvs}
\end{center}
\end{figure}

The surface of each segmented cardiac structure got voxelized to identify its \textbf{surface voxels}. In each surface voxel, the magnitude and the direction of the maximum curvature and the outward normal got calculated. Then, a histogram of the magnitudes of the maximum curvatures got collected across all the surface voxels. \autoref{fig:HistCurvs} shows this histogram for a cardiac surface. In this histogram, the full width at half maximum (FWHM) of the lower peak defined the \textbf{typical curvatures of the myocardium} and the FWHM of the upper peak defined the \textbf{typical curvatures of the cardiac vessels}. This, as shown in \autoref{fig:ModesOfHogsCATs}, was because the myocardium had a larger diameter and thus a lower curvature than the vessels.
Every sample whose patch contained at least one surface voxel was designated to be a \textbf{surface sample}. Among all the surface voxels of every surface sample, median of magnitudes of maximum curvatures, median of directions of maximum curvatures, and median of outward normals got calculated to define the magnitude of maximum curvature, the direction of maximum curvature, and the outward normal of the surface sample, respectively. If the magnitude of maximum curvature of a surface sample was among the typical curvatures of the myocardium/cardiac vessels, then the sample was a myocardium/vessel sample.

To compute the susceptibilities, a search got started from every surface sample by following its outward normal. In each trial of this search, a new sample was hit. If the search was started from a \textbf{myocardium sample}, then the susceptibility of each hit sample $v_{r,j}\in\mathbb{V}_{r,\mathrm{feats}}$ with regard to each class $c\in\mathbb{L}=\{\mathrm{EpAT},\mathrm{PeAT},\mathrm{PvAT},\mathrm{Background}\}$ was defined as
\begin{equation}
s_{r,j,c}=\frac{s'_{r,j,c}}{\sum_{c'\in\mathbb{L}}s'_{r,j,c'}},~~~~s'_{r,j,c}=
\begin{cases}
\text{exp}\Big(\big|\text{cos}(\angle\textbf{d}_{r,j}-\angle\textbf{d}_{\text{myo}})\big|\Big)~~~&\text{if}~c=\text{EpAT}\\
1&\text{otherwise}
\end{cases}.
\label{eq:hitSuscptMyo}
\end{equation}
If the search was started from a \textbf{vessel sample}, then the susceptibility of each hit sample $v_{r,j}\in\mathbb{V}_{r,\mathrm{feats}}$ with regard to each class $c\in\mathbb{L}=\{\mathrm{EpAT},\mathrm{PeAT},\mathrm{PvAT},\mathrm{Background}\}$ was
\begin{equation}
s_{r,j,c}=\frac{s'_{r,j,c}}{\sum_{c'\in\mathbb{L}}s'_{r,j,c'}},~~~~s'_{r,j,c}=
\begin{cases}
\text{exp}\Big(\big|\text{cos}(\angle\textbf{d}_{r,j}-\angle\textbf{d}_{\text{ves}})\big|\Big)&\text{if}~c=\text{PvAT}\\
1&\text{otherwise}\\
\end{cases}.
\label{eq:hitSuscptPerVess}
\end{equation}
Here, $\angle\textbf{d}_{\text{myo}}$ and $\angle\textbf{d}_{\text{ves}}$ denoted the direction of maximum curvature of the myocardium and vessel sample, respectively, and $\angle\textbf{d}_{r,j}$ was the mode of the HOG of the fat channel of the hit sample $v_{r,j}\in\mathbb{V}_{r,\mathrm{feats}}$. This way, every sample $v_{r,j}\in\mathbb{V}_{r,\mathrm{feats}}$ could be hit multiple times or not at all. If it was not hit at all, then its susceptibility $s_{r,j,c}\in(0,1)$ was given by
\begin{equation}
\label{eq:notHitSuscpt}
s_{r,j,c}=\frac{s'_{r,j,c}}{\sum_{c'\in\mathbb{L}}s'_{r,j,c'}},~~~~\forall c\in\mathbb{L}:~s'_{r,j,c}=1.
\end{equation}
If it was hit multiple times, then its final susceptibility with regard to each class was the maximum of all the susceptibilities computed for it with regard to this class. Then the $n_{\mathrm{clas}}=|\mathbb{L}|$ susceptibilities of each sample got normalized to sum to one.

\subsection{Segmentation and Surface Computation of Cardiac Structure}
\label{ssec:SegCardiacStruct}
The cardiac structure was composed of myocardium and cardiac vessels. These lean tissues had high water content and thus appeared bright on water images. Accordingly, to segment them, the water channel of each volumetric fat-water image was enough. The segmentation could be done by computing the voxelwise classification probabilities of the addressed volumetric water image with respect to the cardiac and noncardiac tissues and then finding the class of highest probability in each voxel. This way, voxels whose class of highest probability was the cardiac tissue formed the volume of the cardiac structure.
The voxelwise classification probabilities could be estimated by
\begin{itemize}[leftmargin=*]
\item training the random forest classifier proposed in \cite{Fallah2018a,Fallah2018p,Fallah2019a,FallahJ2019} on some (training) volumetric water images and then processing the addressed volumetric water image with this forest.
\item computing volumetric average and probabilistic atlases of the cardiac structure on some (training) volumetric water images and then registering the addressed image with them.
\end{itemize}

We tried both approaches and observed no significant differences in their performances. However, the computational complexity of the random forest classifier was higher due to its feature extraction, feature selection, and optimization process. This higher complexity come with the provision of additional information including the selected (discriminative) features and the indicators of classification reliabilities of the samples. These additional information were needed for further processing of the samples on our proposed graphs if the initial segmentation was not accurate enough. In contrast to the cardiac adipose tissues, this was not the case for the cardiac structure. Thus, we used the multiatlas registration. To this end, 10 volumetric water images got selected according to the variations in size, age, and gender of volunteers. All of these images become a common size by appending or cropping some zero intensity voxels to their surrounding (normally background) regions.

To compute the average and the probabilistic atlases from the 10 water images, volume of the cardiac structure got manually segmented on each of them. From these segmentations, volumetric binary masks of the same size as the images got formed. In each mask, every voxel whose label was the \textbf{cardiac tissue} had a value of 1 and the rest had a value of 0. Then, one of the water images got randomly selected to be a source (deforming) image and the rest become target (fixed) images. The source image got deformed to the 9 target images through 3D affine transformations applying translation, reflection, scaling, rotation, and shearing by a $4\times 4$ matrix operator. For no transformation, this operator become the identity matrix \cite{Andrei2006}. This way, 10 affinely deformed images (including the source image) and 10 affine transformation fields (including the identity transformation of the source image) were obtained. The intensities of the 10 deformed images got averaged voxelwise to form the \textbf{first average atlas} (corresponding to iteration $i=0$). Then, in each iteration $i\geq 1$ of an iterative nonrigid registration, the $10$ affinely deformed images got further deformed to the average atlas obtained from the previous iteration $i-1$. The intensities of the $10$ deformed images got averaged voxelwise to form the average atlas of the iteration $i$.

\begin{figure}[t!]
\begin{center}
\includegraphics[width=1.0\textwidth]{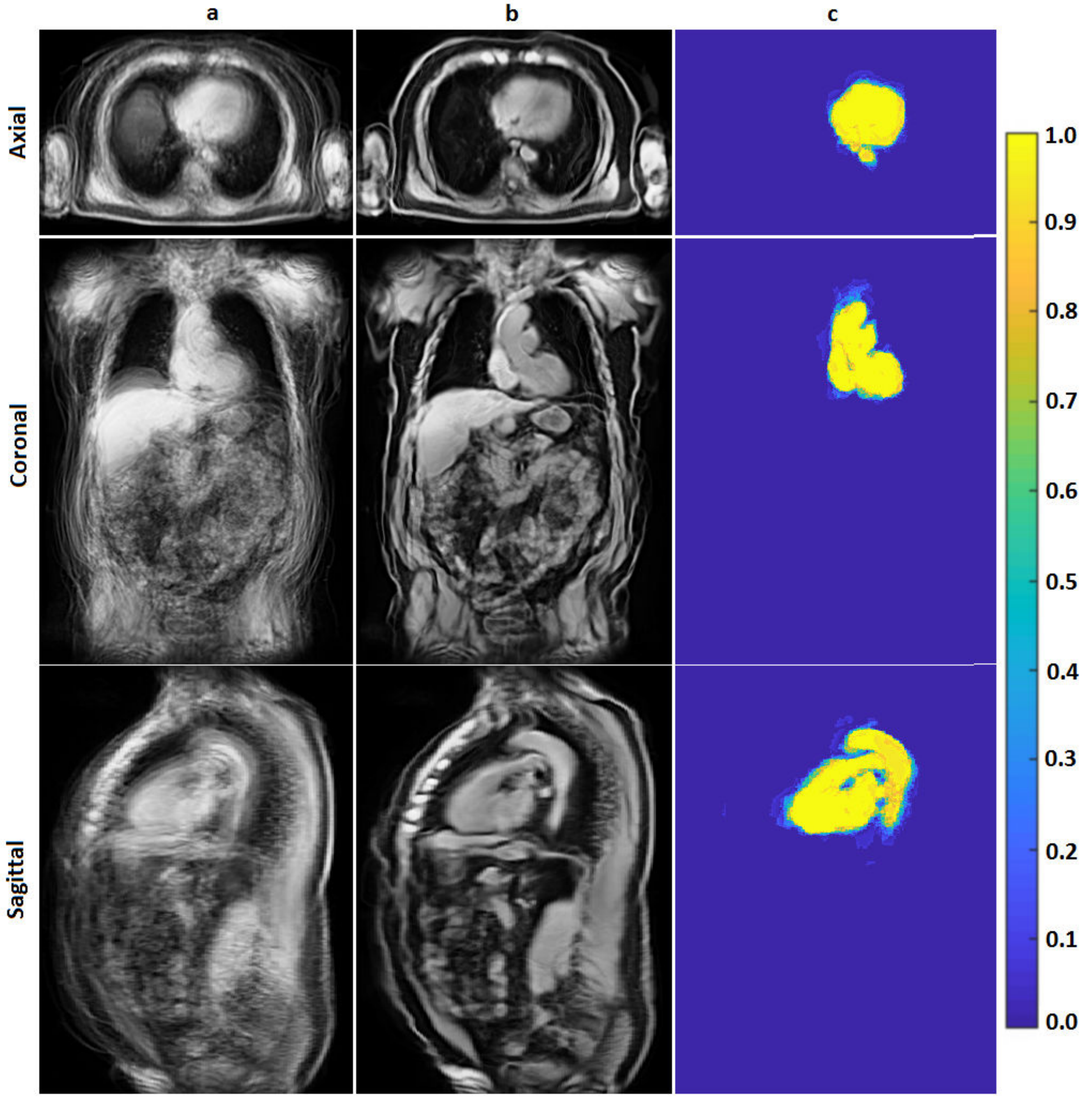}
\caption{Slices of the first average atlas (a), the final average atlas (b), and the probabilistic atlas (c) computed for the cardiac structure on 10 water images.}
\label{fig:AverageProbAtlas}
\end{center}
\end{figure}

\begin{figure}[t!]
\begin{center}
\includegraphics[width=1.0\textwidth]{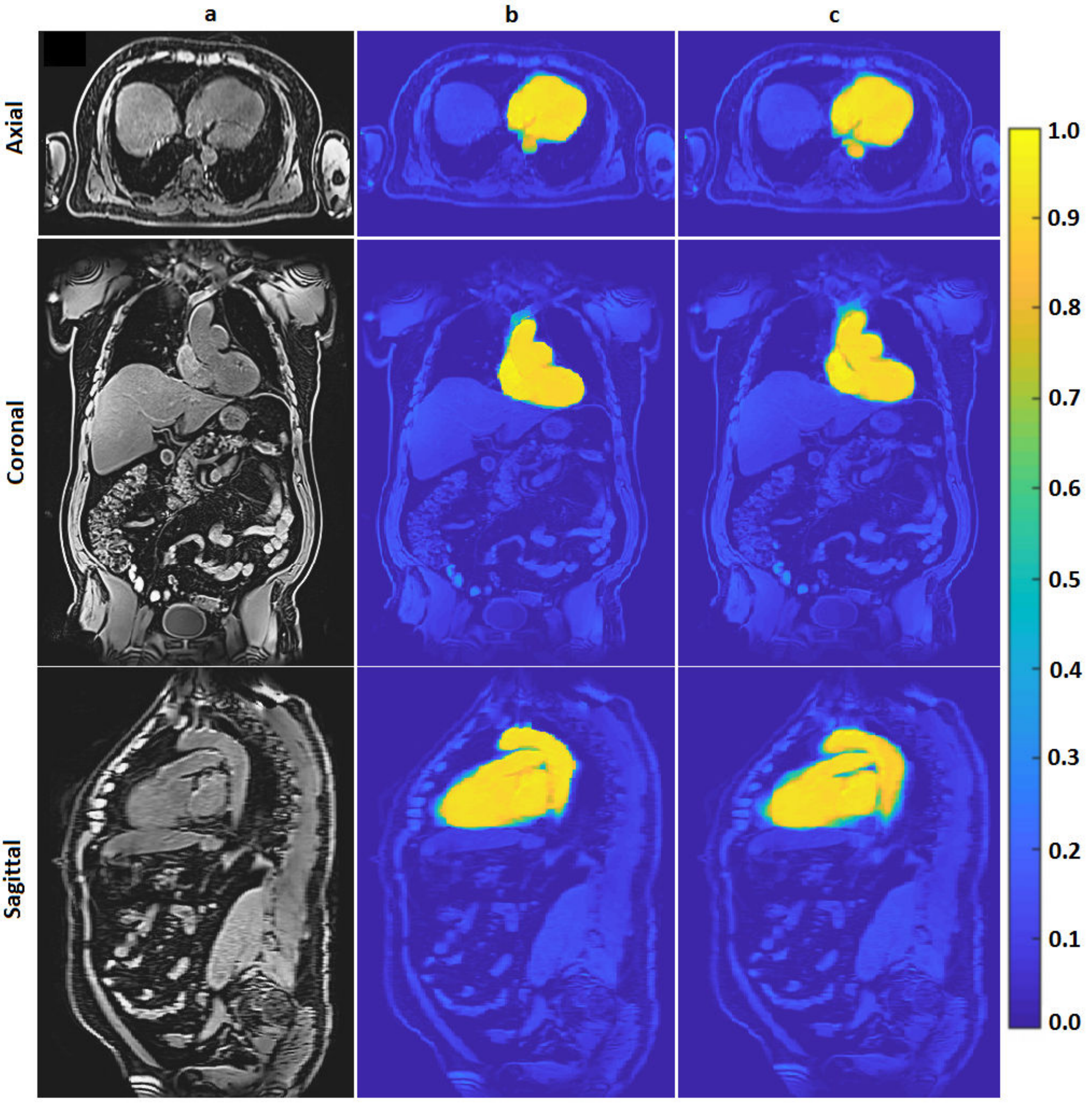}
\caption{The probabilistic maps of the cardiac structure estimated by the multiatlas registration (b) and the random forest classifier proposed in \cite{FallahJ2019} (c) on some slices of a water image (a).}
\label{fig:AtlasRandForestSegCard}
\end{center}
\end{figure}

The nonrigid registrations were based on a free form deformation technique \cite{Rueckert1999}. This technique computed the deformation field by using a B-splines on a grid of control points. To capture local deformations, each grid point was optimized individually. These local optimizations made the registration computationally efficient (even for a large number of control points) and allowed their parallelization for a further gain in efficiency. To speed up the computations, the registrations were done from coarse to fine over multiple resolutions. In each resolution, a certain spacing between the grid points was defined and the deformation field got diffeomorphically accumulated across the iterations \cite{Janssens2011}. The overall deformation field of each resolution become the initial field for the next (finer) resolution. We found $5$ resolution layers and $5$ iterations per resolution to be satisfactory for segmenting the cardiac structure on our volumetric water images. The average atlas obtained after $5$ iterations at the finest resolution was called the \textbf{final average atlas}.

\begin{figure}[t!]
\begin{center}
\includegraphics[width=1.0\textwidth]{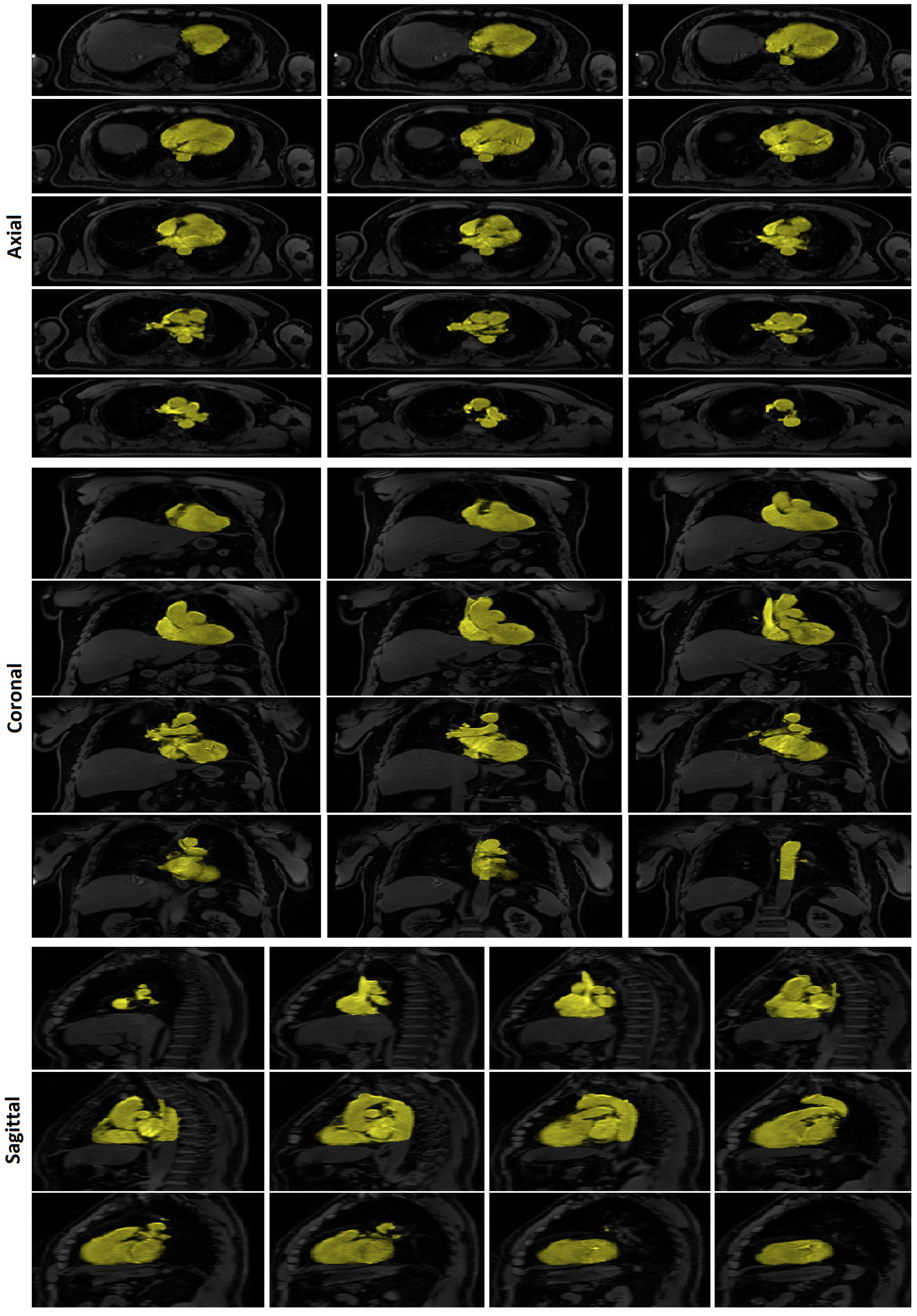}
\caption{A segmented cardiac structure shown on some slices of a volumetric water image.}
\label{fig:SliceSegCardiacStruct}
\end{center}
\end{figure}

For each pair of source-target images, the overall deformation field, resulting from the aforementioned affine and nonrigid registrations, got diffeomorphically accumulated. Then, this field got applied to the volumetric binary mask of the source image. This way, 10 deformed binary masks were obtained. The deformed masks were overlaid and their overlap percentages got counted in every voxel. If $p\%$ of the deformed binary masks had the value of $1$ at a certain voxel, then the probability of this voxel for belonging to the class of \textbf{cardiac tissue} was $p/100$. This volumetric probabilistic map was called the \textbf{probabilistic atlas}.
\autoref{fig:AverageProbAtlas} shows the average and the probabilistic atlases computed for the cardiac structure on 10 water images. By using these computed atlases, the volume of the cardiac structure was segmented on any unseen volumetric water image. To this end, first the final average atlas got registered (deformed) to the unseen image by applying the aforementioned affine and nonrigid registrations. Then, the overall deformation field diffeomorphically accumulated over these registrations got applied to the volumetric probabilistic atlas to form the volumetric probabilistic map of the desired segmentation. \autoref{fig:AtlasRandForestSegCard} compares the probabilistic maps of the cardiac structure estimated by the multiatlas registration and the random forest classifier proposed in \cite{Fallah2018a,Fallah2018p,Fallah2019a,FallahJ2019} on some slices of a volumetric water image. \autoref{fig:SliceSegCardiacStruct} shows a segmented cardiac structure on some slices of a volumetric water image.

After segmenting the volume of the cardiac structure on each volumetric water image, surface of this volume got computed. This surface was a smoothed 3D (iso)surface\footnote{For a continuous function whose domain was the 3D Euclidean space, the isosurface was the set of al 3D points whose values, returned by the function, were a certain constant (isovalue).} returned by an enhanced marching cube algorithm \cite{Custodio2013}. The marching cube algorithm rendered a volume into a 3D isosurface represented by a triangular mesh. That is, it assumed that each piece of the isosurface was a triangle defined by the coordinates of its 3 vertices. Then, it determined the vertices of each triangle by using the cubic patches of the segmented volume. Each patch was represented by its 8 corners' coordinates and had an intensity at each corner. The isovalue determining the isosurface was defined by the user of the algorithm. The marching cube algorithm compared the intensity of each corner of each patch with this isovalue. If the intensity was higher/lower than the isovalue, then the corner got a 1/0 value. A triangular piece of the isosurface intersected an edge of a cubic patch if the patch's corners at the two sides of the edge had different values. The location of the intersection was a vertex of the triangle and was determined according to the differences of the intensities of the edge's corners with the isovalue. That is, the larger the difference of a corner's intensity with the isovalue was, the further away the vertex from the corner would be \cite{Lorensen1987}.

\begin{figure}[t!]
\begin{center}
\includegraphics[width=0.8\textwidth]{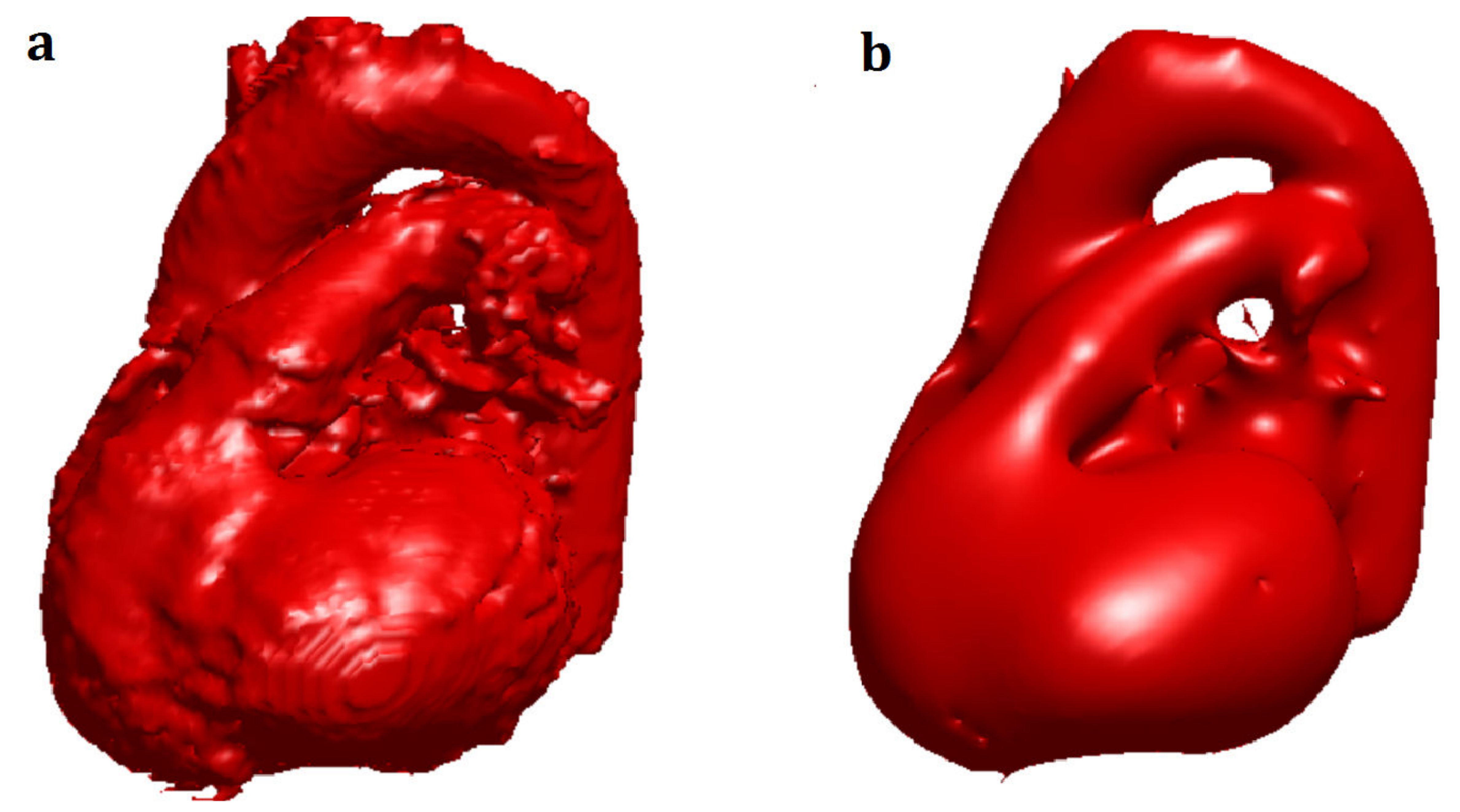}
\caption{Isosurface of a cardiac structure before (a) and after (b) applying the low-pass filter.}
\label{fig:IsoSurfBefAftFlt}
\end{center}
\end{figure}

In our case, the cubic patches of the marching cube algorithm were image patches of at least one voxel belonging to the segmented cardiac volume. To this end, we divided the entire volume of each volumetric water image into nonoverlapping cubic patches of $8$ voxels in a $2\times 2\times 2$ stencil. This way, each patch had 8 voxels as its 8 corners. The coordinates of these voxels were the corner coordinates of the patch and the water intensities of these voxels were compared against the isovalue. We defined the isovalue to be the median of the water intensities of all the voxels forming the segmented cardiac volume.
With 8 corners and 2 values (0 and 1) per corner, each patch had $2^8=256$ possibilities to accommodate a triangular piece of the isosurface. Among these, we could have
\begin{itemize}[leftmargin=*]
\item face ambiguity in each patch whose corners' values were 1 on one diagonal and 0 on the other diagonal and thus could not determine any vertex of the triangular mesh;
\item interior ambiguity in each patch whose corners' values implied multiple triangulations;
\item topological consistence but not correctness of the isosurface. That is, each piece of the isosurface was piecewise linear but the overall isosurface was not due to discontinuities at the patches' boundaries caused by the face or interior ambiguities.
\end{itemize}

These issues were tackled by reducing the possibilities of the triangulations in each cubic patch from $2^8=256$ to 33 cases \cite{Chernyaev1995,Lewiner2003}. In \cite{Custodio2013}, these restrictions got enhanced by additionally producing a topologically correct piecewise linear isosurface. Computation of a topologically correct isosurface implied a higher complexity. The method in \cite{Custodio2013} reduced this complexity by avoiding interpolations at the edges of face or interior ambiguity. The resulting complexity was then linear in the number of triangular pieces (faces) of the triangular mesh of the isosurface. The memory footprint (storage) was also linear in the number of vertices of the isosurface. Thus, we used this method to calculate the triangular mesh of the isosurface of every segmented volume of the cardiac structure. The resulting isosurface was faceted especially at the boundaries of the patches. To smooth it, we applied a linear low-pass filter to it. This filter returned a smoothed triangular mesh after removing high curvature variations of the original mesh without causing any shrinkages \cite{Taubin1995}. \autoref{fig:IsoSurfBefAftFlt} shows the triangular mesh of the isosurface of a segmented cardiac structure before and after applying the low-pass filter.

After smoothing the triangular mesh of each isosurface, its curvatures, introduced in \autoref{ssec:SampClasSuscpt}, got computed. These curvatures were used to define the susceptibilities given by \eqref{eq:hitSuscptMyo}, \eqref{eq:hitSuscptPerVess}, and \eqref{eq:notHitSuscpt} for each sample (vertex) $v_{r,j}\in\mathbb{V}_{r,\mathrm{feats}}$ of each resolution $r\in\{0,\cdots,n_{\mathrm{lay}}\}$. Then, the samples (vertices) $\mathbb{V}_{r,\mathrm{feats}}$ got processed by the neighborhood graph $\mathcal{G}_r=\mathcal{G}_{r,\mathrm{feats}}\bigcup\mathcal{G}_{r,\mathrm{prior}}$ incorporating the SIR model introduced in \autoref{ssec:SirModel}.
It should be noted that the samples could have any resolution (patch size) $r\in\{0,\cdots,n_{\mathrm{lay}}\}$. However, the aforementioned surface computations were only done at the finest resolution, i.e. with respect to the voxels of the segmented cardiac volume.

\begin{figure}[t!]
\begin{center}
\includegraphics[width=0.7\textwidth]{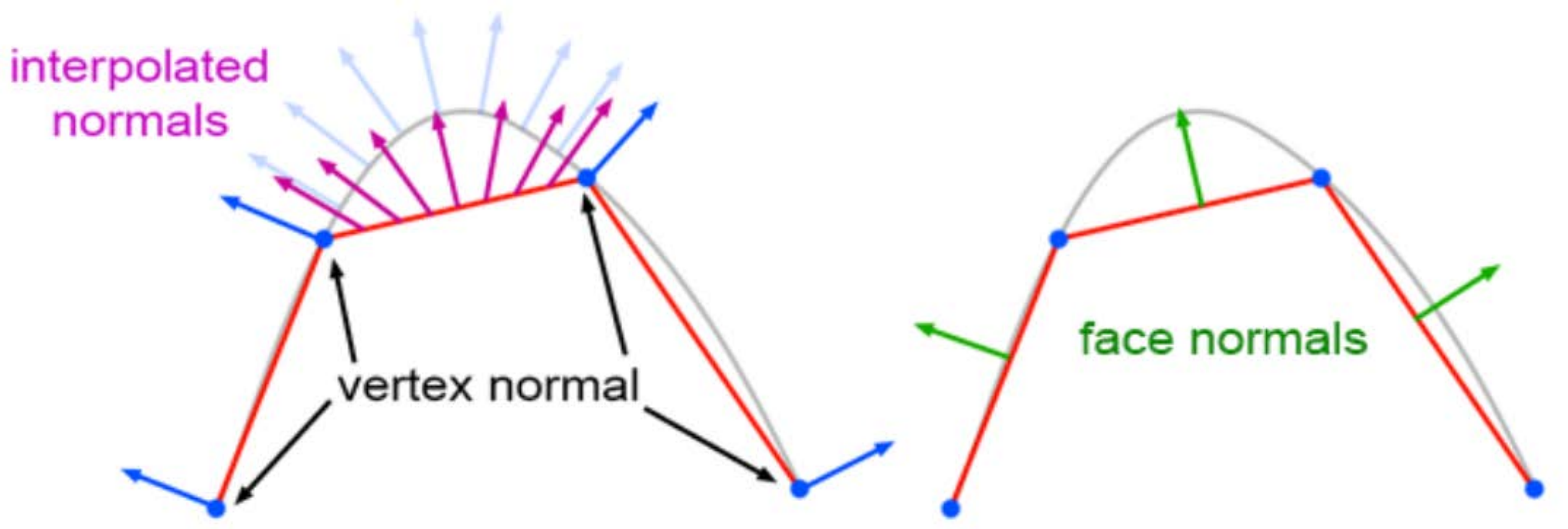}
\caption{Face, vertex, and interpolated normals of a triangular mesh.}
\label{fig:NormalsTriangularMesh}
\end{center}
\end{figure}

As described in \autoref{ssec:SampClasSuscpt}, to compute the minimum, the maximum, the Gaussian, and the mean curvatures of a surface, its curvature tensor\footnote{For a 2D surface like a triangular mesh, the curvature tensor was a matrix.} should be computed. The curvature tensor at every point on a smooth parametrizable surface could be derived from the directional derivatives of the surface normal at the point. That is, the curvature tensor described how the surface normals changed within a neighborhood of the point \cite{Gatzke2006}.
A triangular mesh was not a smooth parametrizable surface rather a piecewise smooth surface. Indeed, each of its triangular pieces (faces) was smooth and parametrizable. This allowed a discrete finite-difference-based scheme to express (approximate) the normal and thus the curvature tensor at every point on a triangular mesh. To this end, first a weighted average of the normals of the triangular faces adjacent to each vertex of the mesh got computed to define the normal at the vertex. The weight of each triangular face was its area. The normal of each triangular face was obtained from the cross product of the vectors aligning with two arbitrarily selected edges of the triangle. To obtain a unit normal, this cross product got divided by its magnitude. The normals at the vertices of the mesh got linearly interpolated to yield the normal at every arbitrary point on the mesh. The linear interpolation of the vertices' normals for obtaining the normal at a point used the interpolation weights required to interpolate the point itself from the vertices. This approach, as depicted in \autoref{fig:NormalsTriangularMesh}, resulted in smoothly varying normals over the triangular mesh.

To compute the directional derivative of the normal at a point on the triangular mesh along a certain direction tangent to the mesh, following steps were taken. First, the direction got represented by a unit vector expressed as a linear combination of two unit vectors aligned with the two edges of the triangular face including the point. Then, along each of these two edges, differences of the normals at the two sides of the edge got computed. The two edge-based normal differences got linearly interpolated to give the desired directional difference of the normal at the point. The weights of this linear interpolation were the weights of expressing the direction as a linear combination of the edges' directions. The curvature tensor at the point got formed from these directional derivatives. A multiplication of this tensor with any vector tangent to the mesh and passing through the point returned the directional derivative of the normal at the point along the vector \cite{Theisel2004,Rusinkiewicz2004}.

\section{Solving Optimizations of Neighborhood Graphs}
\label{sec:solveOptGraphs}
The classification of the samples $\mathbb{D}_r$ on each neighborhood graph $\mathcal{G}_r=\mathcal{G}_{r,\mathrm{feats}}\bigcup\mathcal{G}_{r,\mathrm{prior}}$ proposed in \autoref{sec:FeatPriorGraph}, \autoref{sec:CnstrFeatPriorGraph}, or \autoref{sec:GuidedFeatPriorGraph} resulted in a system of sparse linear equations. This system had a structure like \eqref{eq:linEqsRandWalk2} or \eqref{eq:linEqsUnlab2} depending on whether some of the samples were (manually or automatically) prelabeled. Its coefficients were the elements of the combinatorial Laplace-Beltrami matrix of the graph $\mathcal{G}_r$. Its known information were the prelabeled samples and/or the classification priors of the unlabeled samples. Its solutions were the classification posteriors of the samples.
The combinatorial Laplace-Beltrami matrix was sparse symmetric positive semidefinite and had a dimension of $|\mathbb{D}_r|\times|\mathbb{D}_r|$. For the graphs proposed in \autoref{sec:FeatPriorGraph} and \autoref{sec:CnstrFeatPriorGraph}, it was denoted by $\boldsymbol{\Gamma}_r$ and given by \eqref{eq:combLapMtrx2}. For the graph proposed in \autoref{sec:GuidedFeatPriorGraph}, it was denoted by $\boldsymbol{\Gamma}'_r$ and given by \eqref{eq:combLapMtrx3}. If all the $|\mathbb{D}_r|$ samples could fit into the memory, then each system of equations could be solved through a Cholesky decomposition of $\boldsymbol{\Gamma}_r$ or $\boldsymbol{\Gamma}'_r$. This decomposition was an efficient variant of the lower-upper (LU) decomposition for symmetric positive semidefinite matrices. If all the $|\mathbb{D}_r|$ samples could not fit into the memory, then an iterative preconditioned conjugate gradient optimization could solve \eqref{eq:linEqsRandWalk2} or \eqref{eq:linEqsUnlab2}.

This method was memory efficient and parallelizable on multiple processors. In the absence of noise and disturbances, it provided a monotonically improving approximation of the exact solution in a number of iterations less than $|\mathbb{D}_r|$. Its computational complexity was $\mathcal{O}\big(\mathrm{nnz}(\boldsymbol{\Gamma}_r)\times\sqrt{\kappa(\boldsymbol{\Gamma}_r)}\big)$ or $\mathcal{O}\big(\mathrm{nnz}(\boldsymbol{\Gamma}'_r)\times\sqrt{\kappa(\boldsymbol{\Gamma}'_r)}\big)$ for the system's matrix $\boldsymbol{\Gamma}_r$ or $\boldsymbol{\Gamma}'_r$, respectively, with $\mathrm{nnz}(\cdot)$ denoting the number of nonzero entries and $\kappa(\cdot)$ being the condition number. Thus, the smaller the condition number was, the faster the convergence would be. The inclusion of the aspatial prior-based subgraph $\mathcal{G}_{r,\mathrm{prior}}$ into the graph $\mathcal{G}_r$ added $\lambda_{r,\mathrm{prior}}\in\mathbb{R}_{+}$, introduced in \eqref{eq:weightPriors2}, to every diagonal element of $\boldsymbol{\Gamma}_r$ and $\boldsymbol{\Gamma}'_r$. This in turn reduced $\kappa(\boldsymbol{\Gamma}_r)$ and $\kappa(\boldsymbol{\Gamma}'_r)$ and thereby sped up the convergence of the optimizations. To further speed up solving \eqref{eq:linEqsRandWalk2} or \eqref{eq:linEqsUnlab2}, the preconditioned conjugate gradient method replaced
\begin{equation}
\label{eq:precCondGrad}
\begin{split}
&\boldsymbol{\Gamma}_{r,\mathrm{unl}}\times\mathbf{P}_{r,\mathrm{unl}}+\boldsymbol{\Gamma}_{r,\mathrm{ula}}^T\times\mathbf{P}_{r,\mathrm{lab}}-\lambda_{r,\mathrm{prior}}\times\mathbf{A}_{r,\mathrm{unl}}=\mathbf{0}\\
\text{with}~~~&\mathbf{M}_{\mathrm{pcg}}^{-1}\times\big(\boldsymbol{\Gamma}_{r,\mathrm{unl}}\times\mathbf{P}_{r,\mathrm{unl}}+\boldsymbol{\Gamma}_{r,\mathrm{ula}}^T\times\mathbf{P}_{r,\mathrm{lab}}-\lambda_{r,\mathrm{prior}}\times\mathbf{A}_{r,\mathrm{unl}}\big)=\mathbf{0}\\
\text{or}~~~&\boldsymbol{\Gamma}_r\times\mathbf{P}_r-\lambda_{r,\mathrm{prior}}\times\mathbf{A}_r=\mathbf{0}\\
\text{with}~~~&\mathbf{M}_{\mathrm{pcg}}^{-1}\times\big(\boldsymbol{\Gamma}_r\times\mathbf{P}_r-\lambda_{r,\mathrm{prior}}\times\mathbf{A}_r\big)=\mathbf{0}\\
\text{such~that}~~~&\kappa\big(\mathbf{M}_{\mathrm{pcg}}^{-1}\times\boldsymbol{\Gamma}_r\big)<\kappa(\boldsymbol{\Gamma}_r).
\end{split}
\end{equation}
The preconditioned conjugate gradient method was available through the MATLAB function \textit{pcg}. This function specified the preconditioner matrix $\mathbf{M}_{\mathrm{pcg}}$ and efficiently solved the system
\begin{equation}
\label{eq:precCondSystem}
\begin{split}
&\mathbf{H}_{\mathrm{pcg}}^{-1}\times\boldsymbol{\Gamma}_{r,\mathrm{unl}}\times\mathbf{H}_{\mathrm{pcg}}^{-T}\times\mathbf{Q}_r=\mathbf{H}_{\mathrm{pcg}}^{-1}\times\Big(\lambda_{r,\mathrm{prior}}\times\mathbf{A}_{r,\mathrm{unl}}-\boldsymbol{\Gamma}_{r,\mathrm{ula}}^T\times\mathbf{P}_{r,\mathrm{lab}}\Big)\\
&\text{with}~~~\mathbf{Q}_r=\mathbf{H}_{\mathrm{pcg}}^{T}\times\mathbf{P}_{r,\mathrm{unl}}~~~\text{and}~~~\mathbf{H}_{\mathrm{pcg}}=\mathbf{M}_{\mathrm{pcg}}^{1/2}\\
&\text{or}~~~\mathbf{H}_{\mathrm{pcg}}^{-1}\times\boldsymbol{\Gamma}_r\times\mathbf{H}_{\mathrm{pcg}}^{-T}\times\mathbf{Q}_r=\mathbf{H}_{\mathrm{pcg}}^{-1}\times\Big(\lambda_{r,\mathrm{prior}}\times\mathbf{A}_r\Big)\\
&\text{with}~~~\mathbf{Q}_r=\mathbf{H}_{\mathrm{pcg}}^{T}\times\mathbf{P}_r~~~\text{and}~~~\mathbf{H}_{\mathrm{pcg}}=\mathbf{M}_{\mathrm{pcg}}^{1/2}
\end{split}
\end{equation}
\cite{Barrett1994,Bolz2003}. In all cases, the system's matrix could be $\boldsymbol{\Gamma}'_r$ as well.

\section{Fusion of Posteriors to Labels}
\label{sec:fusePostLabels}
Each variant of the neighborhood graph $\mathcal{G}_r=\mathcal{G}_{r,\mathrm{feats}}\bigcup\mathcal{G}_{r,\mathrm{prior}}$, proposed in \autoref{sec:FeatPriorGraph}, \autoref{sec:CnstrFeatPriorGraph}, or \autoref{sec:GuidedFeatPriorGraph}, estimated the classification posteriors of the samples $\mathbb{D}_r$ of the resolution $r\in\{0,\cdots,n_{\mathrm{lay}}\}$. These posteriors formed a matrix $\mathbf{P}_r={[\mathbf{p}_{r,j}]}_{j,:}={[p_{r,j,c}]}_{j,c}$ of dimension $|\mathbb{D}_r|\times n_{\mathrm{clas}}$. Thus, to estimate the multiresolution posteriors ${\{\mathbf{P}_r\}}_{r=0}^{n_{\mathrm{lay}}}$ of the multiresolution samples ${\{\mathbb{D}_r\}}_{r=0}^{n_{\mathrm{lay}}}$, a stack of multiresolution graphs ${\{\mathcal{G}_r\}}_{r=0}^{n_{\mathrm{lay}}}$ was needed.
These graphs were disconnected from each other and thus acted independently. Hence, they could not incorporate the inter-resolution hierarchical relationships of the multiresolution samples ${\{\mathbb{D}_r\}}_{r=0}^{n_{\mathrm{lay}}}$ into their classifications. To incorporate these relationships into the classifications of the samples ${\{\mathbb{D}_r\}}_{r=0}^{n_{\mathrm{lay}}}$, we used them to \textbf{fuse} the multiresolution posteriors ${\{\mathbf{P}_r\}}_{r=0}^{n_{\mathrm{lay}}}$ into \textbf{hierarchically consistent labels} ${\big\{\mathbf{l}^{*}_r={[l^{*}_{r,j}\in\mathbb{L}]}_j\big\}}_{r=0}^{n_{\mathrm{lay}}}$.
This was achieved by encoding the inter-resolution hierarchical relationships of the multiresolution samples ${\{\mathbb{D}_r\}}_{r=0}^{n_{\mathrm{lay}}}$ into the weighted undirected graph $\mathcal{G}_{\mathrm{hcrf}}=(\mathbb{V}_{\mathrm{hcrf}},\mathbb{E}_{\mathrm{hcrf}})$ of a hierarchical conditional random field (HCRF). That is, the graph $\mathcal{G}_{\mathrm{hcrf}}$ connected the graphs ${\{\mathcal{G}_r\}}_{r=0}^{n_{\mathrm{lay}}}$ with regard to those relationships. Being undirected allowed the graph $\mathcal{G}_{\mathrm{hcrf}}$ to encode the hierarchical relationships in both coarse-to-fine (parent-to-child) and fine-to-coarse (child-to-parent) directions. The graph $\mathcal{G}_{\mathrm{hcrf}}$ processed the multiresolution samples ${\{\mathbb{D}_r\}}_{r=0}^{n_{\mathrm{lay}}}$ by representing them over its vertices $\mathbb{V}_{\mathrm{hcrf}}$. For this processing, all the samples ${\{\mathbb{D}_r\}}_{r=0}^{n_{\mathrm{lay}}}$ should be accompanied with a common set of \textbf{resolution-independent feature types}.
The graph $\mathcal{G}_{\mathrm{hcrf}}=(\mathbb{V}_{\mathrm{hcrf}},\mathbb{E}_{\mathrm{hcrf}})$ had vertices $\mathbb{V}_{\mathrm{hcrf}}=\bigcup_{r=1}^{n_{\mathrm{lay}}}\mathbb{V}_{r,\mathrm{feats}}$ and edges $\mathbb{E}_{\mathrm{hcrf}}=\bigcup_{r=0}^{n_{\mathrm{lay}}-1}\mathbb{E}_{r,r+1}$ with $\mathbb{V}_{r,\mathrm{feats}}$ being the vertices representing the samples $\mathbb{D}_r$ over the spatial feature-based subgraph $\mathcal{G}_{r,\mathrm{feats}}=(\mathbb{V}_{r,\mathrm{feats}},\mathbb{E}_{r,\mathrm{feats}})$ of the neighborhood graph $\mathcal{G}_r=\mathcal{G}_{r,\mathrm{feats}}\bigcup\mathcal{G}_{r,\mathrm{prior}}$. The subset $\mathbb{E}_{r,r+1}$ of the edges was defined as
\begin{equation}
\mathbb{E}_{r,r+1}=\big\{e_{(r,j),(r+1,k)}~|~v_{r,j}\in\mathbb{V}_{r,\mathrm{feats}}~\text{was a child of}~v_{r+1,k}\in\mathbb{V}_{r+1,\mathrm{feats}}\big\}.
\label{eq:hierEdgesHcrf}
\end{equation}
That is, every vertex $v_{r,j}\in\mathbb{V}_{r,\mathrm{feats}}\subset\mathbb{V}_{\mathrm{hcrf}}$ represented a sample and every edge $e_{(r,j),(r+1,k)}\in\mathbb{E}_{r,r+1}\subset\mathbb{E}_{\mathrm{hcrf}}$ represented a hierarchical relationship between $v_{r,j}\in\mathbb{V}_{\mathrm{hcrf}}$ and its parent $v_{r+1,k}\in\mathbb{V}_{r+1,\mathrm{feats}}\subset\mathbb{V}_{\mathrm{hcrf}}$. This way, the graph $\mathcal{G}_{\mathrm{hcrf}}$ connected the graphs $\{\mathcal{G}_r\}_{r=1}^{n_{\mathrm{lay}}}$.

The hierarchical consistency of the estimated labels ${\big\{\mathbf{l}^{*}_r={[l^{*}_{r,j}\in\mathbb{L}]}_j\big\}}_{r=0}^{n_{\mathrm{lay}}}$ implied that the label estimated for each sample should be a major vote of the labels estimated for its hierarchical children in a finer resolution layer. Also, if the label estimated for a sample differed from the label estimated for its hierarchical parent in a coarser resolution layer, then the sample's label should be handled as an outlier. These motivated us to use the outlier detection/removal capability of the Tukey's function introduced in \eqref{eq:TukeyFunc} to derive the edge weights of the graph $\mathcal{G}_{\mathrm{hcrf}}$. However, in contrast to the Tukey's function in \eqref{eq:TukeyFunc}, this Tukey's function did not deal with single-resolution residuals rather residuals derived from features differences of the samples of two subsequent resolutions. Accordingly, for this function, we consolidated the tuning parameters derived for those resolutions into one parameter. The resulting Tukey's function and its tuning parameter were given by
\begin{subequations}
\label{eq:TukeyFunc2}
\begin{align*}
\mathrm{tukey}\big({\|\hat{\mathbf{f}}_{r,j}-\hat{\mathbf{f}}_{r+1,k}\|}_1\big)=&\begin{cases}\frac{\sigma_{r,r+1}^2}{6}\cdot\Big(1-\big[1-(\frac{{\|\hat{\mathbf{f}}_{r,j}-\hat{\mathbf{f}}_{r+1,k}\|}_1}{\sigma_{r,r+1}})^2\big]^3\Big)&\text{if}~{\|\hat{\mathbf{f}}_{r,j}-\hat{\mathbf{f}}_{r+1,k}\|}_1\leq\sigma_{r,r+1}\\
\frac{\sigma_{r,r+1}^2}{6}&\text{otherwise}
\end{cases}\\
\sigma_{r,r+1}=&\min(\sigma_{r,\mathrm{out}},~\sigma_{r+1,\mathrm{out}})\tag{\ref{eq:TukeyFunc2}}
\end{align*}
\end{subequations}
with $\sigma_{r,\mathrm{out}}\in\mathbb{R}_{+}$ being the tuning parameter defined for the resolution $r\in\{0,\cdots,n_{\mathrm{lay}}\}$ in \eqref{eq:resOutRem}, ${\|\cdot\|}_1$ being the $l_1$ norm, and $\hat{\mathbf{f}}_{r,j}$ being the vector of resolution-independent features of the sample represented by the vertex $v_{r,j}\in\mathbb{V}_{r,\mathrm{feats}}\subset\mathbb{V}_{\mathrm{hcrf}}$. Then, our proposed weight $w_{(r,j),(r+1,k)}=w_{(r+1,k),(r,j)}\in\mathbb{R}_{\geq 0}$ for every edge $e_{(r,j),(r+1,k)}\in\mathbb{E}_{r,r+1}\subset\mathbb{E}_{\mathrm{hcrf}}$ was given by
\begin{subequations}
\label{eq:weightHCRF}
\begin{align*}
&w_{(r,j),(r+1,k)}=w_{(r+1,k),(r,j)}=h^{*}_{r,j}~\cdot~h^{*}_{r+1,k}~\cdot~t_{(r,j),(r+1,k)}\tag{\ref{eq:weightHCRF}}\\
&t_{(r,j),(r+1,k)}=t_{(r+1,k),(r,j)}=\begin{cases}\mathrm{exp}\Big(-\mathrm{tukey}\big({\|\hat{\mathbf{f}}_{r,j}-\hat{\mathbf{f}}_{r+1,k}\|}_1\big)\Big)&\text{if}~v_{r,j}~\text{was a child of}~v_{r+1,k}\\0&\text{otherwise}
\end{cases}
\end{align*}
\end{subequations}
with $h^{*}_{r,j}\in(0,1]$ and $h^{*}_{r+1,k}\in(0,1]$ being the indicators of classification reliabilities of $v_{r,j}\in\mathbb{V}_{r,\mathrm{feats}}\subset\mathbb{V}_{\mathrm{hcrf}}$ and $v_{r+1,k}\in\mathbb{V}_{r+1,\mathrm{feats}}\subset\mathbb{V}_{\mathrm{hcrf}}$, respectively. Here, we followed our approach in \eqref{eq:weightTukeyFeats} by multiplying the Tukey's function with the classification reliabilities of the samples in order to account for voxelwise label heterogeneities of the samples' patches.

The HCRF was a conditional random field (CRF) over samples of hierarchical relationships. The CRF was a special case of Markov random field (MRF). Each of these models formulated a classification problem as an \textbf{energy function (sum of single and pairwise potentials)} over a weighted undirected graph representing the samples' relationships. The potentials were formed from the classification posteriors of the samples at their labels. Then, the optimum labels were the minimizers of the energy function. This way, both the MRF and the CRF could fuse the posteriors into the discrete-valued labels by considering the samples' relationships over a graph. However, they differed in the way of expressing the posteriors.
Each sample $v_{r,j}\in\mathbb{V}_{r,\mathrm{feats}}\subset\mathbb{V}_{\mathrm{hcrf}}$ was accompanied with some information listed in page~\pageref{extSmpls}. By denoting these information with $\mathbf{i}_{r,j}$ and using the Bayes' rule, the classification posteriors $\mathbf{p}_{r,j}={[p_{r,j,c}]}_{c\in\mathbb{L}}$ estimated for this sample could be expressed as
\begin{equation}
\label{eq:postDecomp}
\mathbf{p}_{r,j}={\Big[p_{r,j,c}\Big]}_c={\Big[p(c|\mathbf{i}_{r,j})\Big]}_c={\Big[p(\mathbf{i}_{r,j}|c)\cdot p(c)/p(\mathbf{i}_{r,j})\Big]}_c={\Big[p(\mathbf{i}_{r,j},c)/p(\mathbf{i}_{r,j})\Big]}_c
\end{equation}
with $p(\mathbf{i}_{r,j}|c)$ being the likelihood, $p(\mathbf{i}_{r,j},c)$ being the joint probability, and $p(c)$ and $p(\mathbf{i}_{r,j})$ being the prior probability of the class $c\in\mathbb{L}$ and the information $\mathbf{i}_{r,j}$, respectively.

In the MRF, each posterior $p_{r,j,c}=p(c|\mathbf{i}_{r,j})$ was decomposed into the joint and the prior probabilities. Use of the joint probabilistic distribution allowed the MRF to build a \textbf{generative model}. This model was able to both analyze and synthesize the classification information. However, in the CRF, each posterior $p_{r,j,c}=p(c|\mathbf{i}_{r,j})$ was not decomposed into the joint and the prior probabilities. This allowed to only form a \textbf{conditionally discriminative model} of merely analytical capabilities and a less number of parameters than the generative model. For the classification tasks addressed in this dissertation a discriminative model was sufficient. Hence, we used the HCRF and derived the single and pairwise potentials of its energy function from the posteriors ${\{\mathbf{P}_r={[\mathbf{p}_{r,j}]}_{j,:}={[p_{r,j,c}]}_{j,c}\}}_{r=0}^{n_{\mathrm{lay}}}$ of the multiresolution samples ${\{\mathbb{D}_r\}}_{r=0}^{n_{\mathrm{lay}}}$ at their estimated labels ${\big\{\hat{\mathbf{l}}_r={[\hat{l}_{r,j}\in\mathbb{L}]}_j\big\}}_{r=0}^{n_{\mathrm{lay}}}$.

If ${p_{r,j,c}|}_{c=\hat{l}_{r,j}}$ was the posterior of the sample (vertex) $v_{r,j}\in\mathbb{V}_{r,\mathrm{feats}}\subset\mathbb{V}_{\mathrm{hcrf}}$ at the label $\hat{l}_{r,j}\in\mathbb{L}$, then the \textbf{single potential} of this sample was $\mathcal{L}_{r,j}=-\mathrm{log}\big({p_{r,j,c}|}_{c=\hat{l}_{r,j}}\big)$.

If the labels $\hat{l}_{r,j}\in\mathbb{L}$ and $\hat{l}_{r+1,k}\in\mathbb{L}$ were estimated for the samples (vertices) $v_{r,j}\in\mathbb{V}_{r,\mathrm{feats}}\subset\mathbb{V}_{\mathrm{hcrf}}$ and $v_{r+1,k}\in\mathbb{V}_{r+1,\mathrm{feats}}\subset\mathbb{V}_{\mathrm{hcrf}}$, then the \textbf{pairwise potential} of these samples was
\begin{equation}
\label{eq:binPotentHcrf}
\mathcal{L}_{(r,j),(r+1,k)}=w_{(r,j),(r+1,k)}\cdot\delta_{\mathrm{bin}}\big(\hat{l}_{r,j},\hat{l}_{r+1,k}\big)
\end{equation}
with $w_{(r,j),(r+1,k)}\in\mathbb{R}_{\geq 0}$ being the weight given by \eqref{eq:weightHCRF} for the edge $e_{(r,j),(r+1,k)}\in\mathbb{E}_{r,r+1}\subset\mathbb{E}_{\mathrm{hcrf}}$ and $\delta_{\mathrm{bin}}\big(\hat{l}_{r,j},\hat{l}_{r+1,k}\big)\in\{0,1\}$ being the binary distance\footnote{This binary distance was a nonconvex \textbf{distance metric}. Being a distance metric implied $\big[\delta_{\mathrm{bin}}(c,c')=0\iff c=c'\big],~\big[\delta_{\mathrm{bin}}(c,c')=\delta_{\mathrm{bin}}(c',c)\geq 0\big],~\text{and}~\big[\delta_{\mathrm{bin}}(c,c')\leq\delta_{\mathrm{bin}}(c,c'')+\delta_{\mathrm{bin}}(c'',c')\big]$. In contrast, a semi-metric $\delta_{\mathrm{semi}}(c,c')$ fulfilled $\big[\delta_{\mathrm{semi}}(c,c')=0\iff c=c'\big]~\text{and}~\big[\delta_{\mathrm{semi}}(c,c')=\delta_{\mathrm{semi}}(c',c)\geq 0\big]$.} between the labels. That is,
\begin{equation}
\label{eq:binDistMetric}
\delta_{\mathrm{bin}}\big(\hat{l}_{r,j},\hat{l}_{r+1,k}\big)=\begin{cases}
1&\text{if}~~\hat{l}_{r,j}\neq\hat{l}_{r+1,k}\\
0&\text{if}~~\hat{l}_{r,j}=\hat{l}_{r+1,k}
\end{cases}.
\end{equation}

The single and the pairwise potentials formed the energy function of the HCRF as
\begin{equation}
\label{eq:EngHCRF}
\begin{split}
\mathcal{L}_{\mathrm{hcrf}}\Big({\{\hat{\mathbf{l}}_r\}}_{r=0}^{n_{\mathrm{lay}}}\Big)=&\sum_{v_{r,j}\in\mathbb{V}_{\mathrm{hcrf}}}\underbrace{-\mathrm{log}\big({p_{r,j,c}|}_{c=\hat{l}_{r,j}}\big)}_{\mathcal{L}_{r,j}}\\
&+\lambda_{\mathrm{hcrf}}\cdot\sum_{\hspace{-2mm}\substack{e_{(r,j),(r+1,k)}\in\mathbb{E}_{\mathrm{hcrf}}}}\underbrace{w_{(r,j),(r+1,k)}\cdot\delta_{\mathrm{bin}}\big(\hat{l}_{r,j},\hat{l}_{r+1,k}\big)}_{\mathcal{L}_{(r,j),(r+1,k)}}
\end{split}
\end{equation}
with $\lambda_{\mathrm{hcrf}}\in\mathbb{R}_{+}$ being a hyperparameter. Then, the optimum labels were given by
\begin{equation}
\label{eq:optLabelHcrf}
{\big\{\mathbf{l}^{*}_r={[l^{*}_{r,j}\in\mathbb{L}]}_j\big\}}_{r=0}^{n_{\mathrm{lay}}}=\argmin_{{\{\hat{\mathbf{l}}_r\}}_{r=0}^{n_{\mathrm{lay}}}}~\mathcal{L}_{\mathrm{hcrf}}\Big({\{\hat{\mathbf{l}}_r\}}_{r=0}^{n_{\mathrm{lay}}}\Big).
\end{equation}

We conducted the above minimization by using a fast primal-dual algorithm proposed in \cite{Komodakis2008,Zhang2020}. This algorithm had a constant polynomial time complexity. Then, the optimum (hierarchically consistent) labels ${\big\{\mathbf{l}^{*}_r={[l^{*}_{r,j}\in\mathbb{L}]}_j\big\}}_{r=0}^{n_{\mathrm{lay}}}$ estimated for the samples ${\{\mathbb{D}_r\}}_{r=0}^{n_{\mathrm{lay}}}$ got evaluated against their corresponding reference labels ${\big\{\mathbf{l}_r={[l_{r,j}\in\mathbb{L}]}_j\big\}}_{r=0}^{n_{\mathrm{lay}}}$. If $\mathbb{D}_r=\mathbb{D}_{r,\mathrm{val}}$, then these evaluations were used to optimize the hyperparameters of the proposed graphs. If $\mathbb{D}_r=\mathbb{D}_{r,\mathrm{test}}$, then these evaluations were used to measure the overall classification performance of a pipeline consisted of the hierarchical random forest classifier proposed in \cite{Fallah2018a,Fallah2018p,Fallah2019a,FallahJ2019} and the multiresolution graphs proposed in this paper. The test evaluations were done after optimizing the main parameters and the hyperparameters of the forest and the hyperparameters of the graphs.

\section{Graphs' Parameters and Their Optimization}
\label{sec:ParamsGraphs}
The proposed graphs had no trainable (optimizable) parameters. They only involved some tunable hyperparameters and a fixed parameter. The only fixed parameter of the proposed graphs was the number of the resolution layers of our multiresolution pyramid, i.e. $n_{\mathrm{lay}}=5$.

The hyperparameters of the proposed graphs and their discretized values were:
\begin{itemize}[leftmargin=*]
\item parameter controlling the contribution of the priors in estimating the posteriors over the neighborhood graph $\mathcal{G}_r$: $\lambda_{r,\mathrm{prior}}\in\{0.1,0.2,\cdots,0.9,1.0\}$
\item parameter controlling the contribution of the pairwise potentials against the single potentials in the energy function of the HCRF over the graph $\mathcal{G}_{\mathrm{hcrf}}$: $\lambda_{\mathrm{hcrf}}\in\{0.1,0.2,\cdots,0.9,1.0\}$.
\end{itemize}
This way, $\lambda_{r,\mathrm{prior}}\in\mathbb{R}_{+}$ was a hyperparameter of the resolution layer $r\in\{0,\cdots,n_{\mathrm{lay}}\}$ and $\lambda_{\mathrm{hcrf}}\in\mathbb{R}_{+}$ was a resolution-independent hyperparameter.
These hyperparameters got tuned for automatically segmenting $n_{\mathrm{clas}}=|\mathbb{L}|=8$ classes of vertebral bodies (VBs), intervertebral disks (IVDs), psoas major (PM) and quadratus lumborum (QL) muscles, epicardial adipose tissues (EpAT), pericardial adipose tissues (PeAT), cardiac perivascular adipose tissues (PvAT), and background on each volumetric fat-water image. The parameters ${\{\lambda_{r,\mathrm{prior}}\in\mathbb{R}_{+}\}}_{r=0}^{n_{\mathrm{lay}}}$ got optimized from the coarsest resolution $r=n_{\mathrm{lay}}$ to the finest resolution $r=0$. To this end, in each resolution layer $r\in\{0,\cdots,n_{\mathrm{lay}}\}$, several optimization trials got conducted. In each trial, a value for $\lambda_{r,\mathrm{prior}}\in\mathbb{R}_{+}$ got randomly selected from the set of discretized values. With this value, the neighborhood graph $\mathcal{G}_r$ which could be one of the graphs proposed in \autoref{sec:FeatPriorGraph}, \autoref{sec:CnstrFeatPriorGraph}, or \autoref{sec:GuidedFeatPriorGraph} got built. Then, the classification performance of this graph got evaluated on the validation samples $\mathbb{D}_{r,\mathrm{val}}$. In this process, each neighborhood graph $\mathcal{G}_r$ computed the classification posteriors $\mathbf{P}_r={[\mathbf{p}_{r,c}]}_{:,c}={[\mathbf{p}_{r,j}]}_{j,:}={[p_{r,j,c}]}_{j,c}$ of the samples $\mathbb{D}_{r,\mathrm{val}}$ by using their information listed in page~\pageref{extSmpls}. For each sample $v_{r,j}\in\mathbb{D}_{r,\mathrm{val}}$, the estimated label was $\hat{l}_{r,j}=\argmax_{c}~p_{r,j,c}$. The estimated labels $\hat{\mathbf{l}}_r={[\hat{l}_{r,j}\in\mathbb{L}]}_j$ got evaluated against their corresponding reference labels $\mathbf{l}_r={[l_{r,j}\in\mathbb{L}]}_j$ by calculating the \textit{precision} and the \textit{recall} metrics for each of the $n_{\mathrm{clas}}-1=7$ foreground classes against the rest of the classes. This way, 7 \textit{precision} and 7 \textit{recall} values were obtained. These values got averaged to represent the overall classification performance of the neighborhood graph. The optimization trials continued by randomly selecting another value for $\lambda_{r,\mathrm{prior}}\in\mathbb{R}_{+}$ until the resulting graph could not exceed the averaged \textit{precision} and \textit{recall} values of any of the graphs of the same kind in the last 20 trials.

\begin{table}[t!]
\begin{center}
\caption{Optimized hyperparameters and the overall time of optimizing the proposed graphs.}
\label{table:hypOptGraphs}
\vspace{2mm}
\resizebox{0.95\textwidth}{!}{%
\begin{tabular}{|c|ccccccc|c|}
\hline
&\multicolumn{7}{c|}{\textbf{Hyperparameters}}&\textbf{Optimization}\\
\textbf{Graph}&$\lambda_{0,\mathrm{prior}}$&$\lambda_{1,\mathrm{prior}}$&$\lambda_{2,\mathrm{prior}}$&$\lambda_{3,\mathrm{prior}}$&$\lambda_{4,\mathrm{prior}}$&$\lambda_{5,\mathrm{prior}}$&$\lambda_{\mathrm{hcrf}}$&\textbf{Time}\\\hline
FPG$^{1}$&1.0&$0.9$&$0.8$&$0.7$&$0.7$&$0.5$&$0.9$&132 mins\\
CFPG$^{1}$&0.8&$0.6$&$0.5$&$0.5$&$0.4$&$0.3$&$0.7$&121 mins\\
GFPG$^{2}$&0.9&$0.7$&$0.6$&$0.5$&$0.4$&$0.3$&$0.8$&143 mins\\\hline
\multicolumn{9}{l}{\small FPG: feature- and prior-based graph proposed in \autoref{sec:FeatPriorGraph}}\\
\multicolumn{9}{l}{\small CFPG: constrained feature- and prior-based graph proposed in \autoref{sec:CnstrFeatPriorGraph}}\\
\multicolumn{9}{l}{\small GFPG: guided feature- and prior-based graph proposed in \autoref{sec:GuidedFeatPriorGraph}}\\
\multicolumn{9}{l}{\small 1: segmenting VBs, IVDs, PM and QL muscles, EpAT, PeAT, PvAT, and background}\\
\multicolumn{9}{l}{\small 2: segmenting cardiac adipose tissues (EpAT, PeAT, and PvAT) and background}
\end{tabular}}
\end{center}
\end{table}

Similar optimization trials got conducted on the graph $\mathcal{G}_{\mathrm{hcrf}}$ to find the best value for $\lambda_{\mathrm{hcrf}}\in\mathbb{R}_{+}$. These optimizations were done for best performing neighborhood graphs ${\{\mathcal{G}_r\}}_{r=0}^{n_{\mathrm{lay}}}$ of each kind obtained from the above process. More specifically, for each kind of the neighborhood graphs ${\{\mathcal{G}_r\}}_{r=0}^{n_{\mathrm{lay}}}$ which could be a kind proposed in \autoref{sec:FeatPriorGraph}, \autoref{sec:CnstrFeatPriorGraph}, or \autoref{sec:GuidedFeatPriorGraph}, several optimization trials got conducted to find the best value of $\lambda_{\mathrm{hcrf}}\in\mathbb{R}_{+}$. In each trial, a value for $\lambda_{\mathrm{hcrf}}\in\mathbb{R}_{+}$ got randomly selected from the set of discretized values. With this value, the hierarchical graph $\mathcal{G}_{\mathrm{hcrf}}$ got built. Then, over this graph, the multiresolution posteriors ${\{\mathbf{P}_r\}}_{r=0}^{n_{\mathrm{lay}}}$ obtained from the best performing neighborhood graphs ${\{\mathcal{G}_r\}}_{r=0}^{n_{\mathrm{lay}}}$ got fused into the hierarchically consistent labels ${\big\{\mathbf{l}^{*}_r={[l^{*}_{r,j}\in\mathbb{L}]}_j\big\}}_{r=0}^{n_{\mathrm{lay}}}$. These labels got then evaluated against their corresponding reference labels ${\big\{\mathbf{l}_r={[l_{r,j}\in\mathbb{L}]}_j\big\}}_{r=0}^{n_{\mathrm{lay}}}$ by calculating the \textit{precision} and the \textit{recall} metrics for each of the $n_{\mathrm{clas}}-1=7$ foreground classes against the rest of the classes. This way, 7 \textit{precision} and 7 \textit{recall} values were obtained. These values got averaged to represent the overall performance of the hierarchical graph $\mathcal{G}_{\mathrm{hcrf}}$. The optimization trials continued by randomly selecting another value for $\lambda_{\mathrm{hcrf}}\in\mathbb{R}_{+}$ until the resulting graph could not exceed the averaged \textit{precision} and \textit{recall} values of any of the graphs in the last 20 trials. The \textit{precision} and \textit{recall} metrics were selected due to their robustness against the imbalanced class-sample distributions. \autoref{table:hypOptGraphs} shows the optimized hyperparameters and the overall time of optimizing them for each kind of the neighborhood graphs ${\{\mathcal{G}_r\}}_{r=0}^{n_{\mathrm{lay}}}$ on a PC with 16 GB RAM and a quad-core CPU of 3.10 GHz frequency.

Among the types of the neighborhood graphs, the guided feature- and prior-based graph proposed in \autoref{sec:GuidedFeatPriorGraph} required to segment additional tissues adjacent to the addressed tissues in order to extract additional information for guiding the segmentations. These additional segmentations become cumbersome when the addressed tissues had separate parts or were surrounded by tissues of high anatomical and contrast variations. Examples of these tissues were VBs, IVDs, and muscles. Thus, we used the guided feature- and prior-based graph only to segment the cardiac adipose tissues.

As shown in \autoref{table:hypOptGraphs}, in each resolution $r\in\{0,\cdots,n_{\mathrm{lay}}\}$, the priors' weight $\lambda_{r,\mathrm{prior}}\in\mathbb{R}_{+}$ of the constrained feature- and prior-based graph was lower than this parameter for the other graphs. This could be attributed to downweighting the roles of the priors in the classifications of the constrained graph due to absorbing samples of highly confident priors into the prelabeled sets and only keeping samples of low confident priors for the classifications. Also, in each resolution, the priors' weights of the feature- and prior-based graph were slightly higher than those of the guided feature- and prior-based graph. This could be attributed to the use of additional information in the guided graph which in turn reduced the roles of the priors. Furthermore, in all the graph variants, the priors' weights got decreased by increasing the resolution (reducing $r$). This revealed that the priors contributed more to tissue localizations in the coarser resolution layers than to accurate segmentations in the finer resolution layers. Finally, the hyperparameter $\lambda_{\mathrm{hcrf}}\in\mathbb{R}_{+}$ reached its minimum in the constrained feature- and prior-based graph. This could be because it was hard to propagate the detected boundaries and the estimated classifications from one resolution to another resolution in the challenging segmentation regions of low confident priors.

\begin{singlespace}
{\footnotesize
\bibliography{ArtklBookMisc,IEEEabrvIndexMedicus,ProcLong}}
\end{singlespace}
\end{document}